\documentclass[11pt]{article}
\usepackage{jheppub}
\usepackage{amssymb,amsfonts}
\usepackage{mathrsfs}
\usepackage{subcaption}
\usepackage{graphicx}
\usepackage{booktabs}
\let\savenumberline\numberline
\def\numberline#1{\savenumberline{#1.}}
\makeatletter
\renewcommand{\@seccntformat}[1]{\csname the#1\endcsname.\,\,}
\makeatother
\newcommand\secref[1]{{\S\ref{#1}}}
\newcommand\appref[1]{{Appendix~\ref{#1}}}
\newcommand\figref[1]{{Figure~\ref{#1}}}
\newcommand\tabref[1]{{Table~\ref{#1}}}

\newcommand{\CD}{{\cal D}}

\newcommand{\CN}{{\cal N}}

\newcommand{\MR}{{\mathbb R}}

\newcommand{\EV}{\mathbb E}

\newcommand{\Xt}{X(t)}
\newcommand{\Yt}{Y(t)}
\renewcommand{\tilde}[1]{\widetilde{#1}}
\renewcommand{\hat}[1]{\widehat{#1}}
\newcommand{\be}{\begin{equation}}
\newcommand{\ee}{\end{equation}}
\newcommand{\bea}{\begin{eqnarray}}
\newcommand{\eea}{\end{eqnarray}}

\makeatletter
\def\@fpheader{\relax}
\makeatother
\usepackage{graphicx}
\usepackage{latexsym}

\title{Dynamical Regimes of Multimodal Diffusion Models}
\author[a,b,1]{Emil Albrychiewicz,}
\author[a,b,c,1]{Andr\'{e}s Franco Valiente,}
\author[d]{and Li-Ching Chen}
\affiliation[a]{Leinweber Institute for Theoretical Physics and Department of Physics,\\
University of California, Berkeley, CA, 94720-7300, USA}

\affiliation[b]{Theoretical Physics Group, Lawrence Berkeley National Laboratory,\\
Berkeley, CA 94720-8162, USA}

\affiliation[c]{Department of Radiation Oncology, University of California, San Francisco}

\affiliation[d]{
Computational Precision Health, University of California San Francisco, 2177 Hearst Ave, Berkeley, CA, 94709, USA
 }

\affiliation[1]{These authors contributed equally.}

\emailAdd{ealbrych@berkeley.edu}
\emailAdd{andresfranco@berkeley.edu}
\emailAdd{liching\_chen@berkeley.edu}

\abstract{
Diffusion based generative models have achieved unprecedented fidelity in synthesizing high dimensional data, yet the theoretical mechanisms governing multimodal generation remain poorly understood. Here, we present a theoretical framework for coupled diffusion models, using coupled Ornstein-Uhlenbeck processes as a tractable model. By using the nonequilibrium statistical physics of dynamical phase transitions, we demonstrate that multimodal generation is governed by a spectral hierarchy of interaction timescales rather than simultaneous resolution. A key prediction is the ``synchronization gap'', a temporal window during the reverse generative process where distinct eigenmodes stabilize at different rates, providing a theoretical explanation for common desynchronization artifacts. We derive analytical conditions for speciation and collapse times under both symmetric and anisotropic coupling regimes, establishing strict bounds for coupling strength to avoid unstable symmetry breaking. We show that the coupling strength acts as a  spectral filter that enforces a tunable temporal hierarchy on generation. We support these predictions through controlled experiments with diffusion models trained on MNIST datasets and exact score samplers. These results motivate time dependent coupling schedules that target mode specific timescales, offering a potential alternative to ad hoc guidance tuning.
}

\begin{document}

\maketitle

\section{Introduction}
The recent success of diffusion based generative models has fundamentally transformed the landscape of unsupervised learning, enabling the synthesis of high dimensional data with unprecedented fidelity across modalities such as image, audio, and video \cite{sohl2015deep, song2019generative, song2020denoising, song2020score, kong2020diffwave, chen2020wavegrad, saharia2022photorealistic, ho2022videodiffusion, ho2022imagenvideo, poole2022dreamfusion, yang2023diffusion, peebles2023scalable, lai2025principles, DeepLearning24, pml2Book}. While the empirical performance of these models is well documented, the theoretical mechanisms governing the dynamics of the reverse-time generative process remain an active area of inquiry \cite{ambrogioni2023statistical, kadkhodaie2023generalization, yoon2023diffusion, cui2023analysis}. Recent work by Biroli et al. (2024) \cite{biroli2024} provides a rigorous statistical mechanics framework for understanding these dynamics in single-variable models, delineating three distinct dynamical regimes: (I) a noise regime, dominated by high-entropy disorder; (II) a speciation regime, where the generative trajectory converges toward the data manifold and samples from the underlying population distribution; and (III) a collapse regime, characterized by memorization where the process concentrates onto the discrete empirical distribution of the training data. This dynamical description of the denoising process provides theoretical context for an important question when diffusion models generate new samples versus recreating memorized ones. 

However, the extension of this framework to multimodal generation introduces nontrivial theoretical complexities, e.g., text-to-image, audio-video problems. Multimodality imposes additional degrees of freedom and necessitates the satisfaction of two distinct constraints: intra-modal content coherence i.e. the structural integrity of each modality, and cross modal alignment \cite{radford2021learning, Rombach2021-li, saharia2022photorealistic} i.e. the semantic synchronization between modalities. In this work, we address these constraints within the framework of coupled diffusion models. The coupled diffusion framework acts as the minimal candidate for a universality class for multimodal generation, explicitly representing the statistical dependencies between modalities as an interaction term that constrains the joint probability statistical manifold. We argue using the theoretical model that these constraints do not necessarily resolve simultaneously. We demonstrate that the coupling between modalities induces distinct convergence scales, creating a hierarchy of transition times where content formation and alignment stabilize at different stages of the reverse process.

To formalize these dynamics, we introduce a framework based on coupled Ornstein-Uhlenbeck (OU) processes \cite{uhlenbeck1930theory}. Consider a stochastic process driven purely by a Brownian motion, the variance would grow linearly with time, causing signal values to diverge to infinity and leading to numerical instability. In contrast, the OU process is characterized by a mean-reverting drift term that effectively dampens noise injection. This mean inversion ensures that the process remains within a bounded, numerically stable range and converges to a stationary distribution under specific conditions. Therefore, we choose to model the diffusion architecture via an OU process rather than a simple Brownian motion. Crucially, the existence of this stationary distribution provides a well defined, closed form prior, $\mathcal{N}(0, I)$, from which the reverse generation process can be initialized. Our approach aligns with the variance preserving formulation proposed by Song et al. (2021) \cite{song2020score}, which carefully balances the decay rate and noise magnitude to ensure the total variance of the signal remains constant throughout the forward dynamics.

In this work, we provide a comprehensive theoretical analysis of coupled diffusion models in \secref{sec:theory}. We extend work of \cite{biroli2024} to apply statistical physics models, concretely random energy model (REM) \cite{derrida1981random}, to study the dynamical regimes of coupled diffusion processes. Our findings are the following. We derive a generalized speciation time argument as bifurcation point where deterministic drift field acquires additional fixed points \eqref{eqn:SpeciationTime} and review REM arguments for derivation of collapse time condition \eqref{eqn:CritCond}. Throughout this paper, we adopt the nomenclature of \cite{biroli2024} and we refer to the transition time between regime I and regime II as a \textit{speciation time} $t_S$, for a transition between regime II and regime III we call this time a \textit{collapse time} $t_C$.  

In \secref{sec:IsoCoup}, we derive the analytical solution for the reverse process under symmetric linear coupling. By diagonalizing the system into common and difference eigenmodes, we show that speciation time condition can be written as a sum of signal to noise ratios (SNRs) for decoupled modes \eqref{eqn:SpecSumSNRs}. Due to different eigenvalues these modes lead to a separation of time scales. We call this effect as \textit{synchronization gap}, a temporal window during the reverse process when one eigenmode crossed regime but the other has not. This gap exists for both speciation and collapse time and its size depends on the coupling strength $g$ as can be seen on \figref{fig:SpeciationTime} for speciation time and \figref{fig:CollapseTime} for collapse time. We suggest that this gap can be used to explain desynchronization artifacts often observed in multi modal generation. The coupling strength $g$ acts as a spectral filter that enforces a temporal hierarchy on generation, ensuring that shared semantic structures emerge significantly earlier than modality-specific discrepancies. 

The transcendental equation for collapse time is given in \eqref{eqn:ColTimeEqn}. We confirm the curse of dimensionality findings of \cite{biroli2024} i.e. if number of samples does not scale exponentially with sample dimension, the collapse time cannot be found. For a chosen set of parameters, the collapse time remains nearly constant when coupling strength $g$ is increased in difference to speciation time that changes significantly. We also derive a strict bound for a coupling strength $g$ \eqref{eqn:InvCond} beyond which the model undergoes unstable symmetry breaking \cite{biroli2023generative, raya2023spontaneous, yu2025nonequilbrium} and sampling trajectories diverge. A cumulant generating function for computation of collapse time is derive in \appref{app:DerCGF}. 

In \secref{sec:AsymCoup}, we introduce an anisotropic coupling matrix, which is relevant for conditional generation tasks. In this case,  information is flowing from one modality to another without a reciprocal feedback. We derive equations for speciation and collapse times and solve them numerically for particular choices of initial distributions. The closed form expressions are listed in \appref{app:ExpCt}. As in the symmetric case, the speciation time is more dependent on the coupling strength \figref{fig:SpecTimeAsym} whereas collapse time does not change much \figref{fig:AsymCollapse}. We plot the dependence of speciation time on the alignment angle between condition and target means in \figref{fig:PhaseDiag}, showing that large coupling can remove bifurcation points for aligned directions whereas the effect on misaligned directions is much smaller. We also show that conditional collapse time can push collapse time further, demonstrating that external guidance delays the transition to the collapse regime. Importantly, the collapse time does not depend on the angle between condition and target means. 

To provide evidence for the theory discussed in the paper, we conduct two experiments that we describe in \secref{sec:MinstExp} and \appref{sec:toy_ou_experiment}. In the former, we train a diffusion model on the MNIST image dataset \cite{lecun2002gradient} and we use deterministic Denoising Diffusion Implicit Model (DDIM) \cite{song2020denoising} and stochastic Denoising Diffusion Probabilistic Model (DDPM) \cite{ho2020denoising}. This model is supposed to act as a proxy for a symmetric coupled OU discussed in \secref{sec:IsoCoup}. Although we do not change the architecture of the model by explicitly introducing the coupling, we construct pairs of samples that we diagonalize using common and difference eigenmodes. The coupling strength modifies diffusion noise covariance. We test whether such model exhibits a synchronization gap and how it changes when coupling strength is adjusted. First, using the experiment with DDIM sampler. We confirm that the synchronization gap cannot be attributed to fixed channel conventions or marginal statistics of the dataset \tabref{tab:synch_gap}. We also provide an evidence of a desynchronization effect \figref{fig:MainGhostTraj} and show that noise injection in the window when one mode stabilized and the other not affects the not yet stabilized mode more \figref{fig:InterventionPlots}. Since the speciation time is a point corresponding to branching of stochastic trajectories we employ DDPM sampler to study cloning samples and observing their trajectories in the reverse time. By training an image classifier we can determine when clones agree at the final time of denoising process. With that metric one can determine whether the denoising process crossed from the regime I to regime II. Indeed we find that that the synchronization gap appears when the coupling is introduced and its width depends on the coupling strength size \figref{fig:SpeciationSweep}. 

In the latter experiment \appref{sec:toy_ou_experiment} we design a conditional generation corresponding to anisotropic coupling discussed in \secref{sec:AsymCoup}. In this case we do not train a neural network, instead we use reverse time sampler with an exact score. We test different values of coupling, applied with different schedules and we find that coupling effect is angle dependent as expected from the theory. We report that small to moderate coupling is beneficial for misaligned modalities. However, constant coupling schedule has a detrimental effect for aligned modalities whereas late coupling has the least detrimental effect on the aligned modalities yet it ameliorates misaligned ones. This exact experiment provides an intuition for constructing coupling schedules in realistic examples. 

In summary, we believe this framework advances the theoretical understanding of diffusion models by establishing a direct link between cross modal interaction mechanisms and nonequilibrium phase transitions. By quantifying the synchronization gap, we provide a basis for moving beyond the heuristic tuning of guidance strengths toward the principled design of coupling schedules that respect the intrinsic timescales of semantic emergence across modalities. This perspective suggests that future multimodal diffusion architectures should be approached not merely as engineering black boxes, but as nonequilibrium statistical systems whose stability, convergence speed and quality is governed by the spectral hierarchy of their interactions.

\section{Theoretical Framework of Coupled Diffusion Models}
\label{sec:theory}
For the following sections, we will use uppercase for random variables and lowercase variables for their corresponding realizations. 

We begin by modeling a coupled diffusion model architecture via a coupled system of two $d-$dimensional Ornstein-Uhlenbeck stochastic processes  $Z(t)= (X(t),Y(t)) \in \mathbb{R}^{2d}$ described by the Itô stochastic differential equation
\begin{align}
\label{eqn:SDE}
    dZ(t)=MZ(t)\, dt+\Sigma_W\,dW(t),
\end{align}
driven by a $\mathbb{R}^{2d}$ valued Wiener process $W(t)$ with volatility matrix $\Sigma_W$. For theoretical clarity, we keep dimensions of $\Xt \in \mathbb{R}^d$ and $\Yt \in \mathbb{R}^d$ to be equal although typically the embedding spaces for different modes (e.g. image and video) have different dimensionality in practice.  In what follows, we will discuss two different choices of the relaxation matrix $M$, which we set to be time independent. The intuition behind either choice will be how we wish to model the information flow, we will begin by investigating the case of a symmetric relaxation matrix where there is no preferred information flow between either modality. In the second case, we will break this symmetry and introduce an anisotropic relaxation matrix where there is a clear directionality between the information flow between the two modalities.

First, we discuss the case of an symmetric relaxation matrix with couplings $g$ and $\beta$ i.e.
\begin{align}
\label{eqn:SymMandS}
    M=\begin{pmatrix}
    -\beta & g \\
    g & -\beta 
    \end{pmatrix}\otimes I_d, \qquad
    \Sigma_W=\begin{pmatrix}
    \sigma_W & 0 \\
    0 & \sigma_W
    \end{pmatrix}\otimes I_d,
\end{align}
with a stability condition $\beta > |g| $ that ensures the eigenvalues have a negative real part. $I_d$ is the d dimensional identity matrix.

The case of an anisotropic relaxation matrix will be the choice of a lower triangular relaxation matrix where we explicitly have
\begin{align}
\label{eqn:AsymMandS}
    M=\begin{pmatrix}
    -\beta & 0 \\
    g & -\beta
    \end{pmatrix}\otimes I_d, \qquad
    \Sigma_W=\begin{pmatrix}
    \sigma_{W} & 0 \\
    0 & \sigma_{W}
    \end{pmatrix}\otimes I_d,
\end{align}
in this case there is no stability restriction on the value of $g$ coupling provided that $\beta>0$. We note that this lower triangular case has an explicit decomposition where it can be written as a sum of a diagonal matrix $\beta I_2$ and a nilpotent matrix
\begin{equation}
M=-\beta I_2 \otimes I_d+\left(\begin{array}{ll}
0 & 0 \\g & 0
\end{array}\right) \otimes I_d.
\end{equation}
This linear system of SDEs can be explicitly solved through the use of Itô's lemma. The solution is described by a Gaussian process. In order to see this, we begin by writing the SDE as
\begin{align}
    d(e^{-Mt}Z(t))=e^{-Mt}\Sigma_W dW(t),
\end{align}
which can be integrated, given initial conditions $Z(0)$ to give the Gaussian process
\begin{align}
    Z(t)=e^{Mt}Z(0)+e^{Mt}\int_0^te^{-Ms}\Sigma_WdW(s).
\end{align}
Note that in the lower triangular anisotropic case, the matrix exponential simplifies due to the nilpotent part of $M$ giving us the useful identity $e^{-Ms}=e^{-\beta(-s)}(I-Ns)$. 

From this solution, we can then calculate correlation functions such as the mean
\begin{align}
\label{eqn:MeanDrift}
    \mu(t)&=\EV[Z(t)]=e^{Mt}\mu(0),
\end{align}
and the covariance matrix
\begin{align}
\label{eqn:CondCov}
    Q(t)&=\EV\left[(Z(t)-\mu(t))(Z(t)-\mu(t))^\intercal\right] \\
    &=\int_0^te^{M\tau}\Sigma_W \Sigma_W^T e^{M^\intercal\tau}d\tau.
\end{align}
For the stochastic process described by \eqref{eqn:SDE} to admit a well-defined transition probability density with respect to the Lebesgue measure on $\mathbb{R}^{2d}$, the covariance matrix $Q(t)$ must be invertible and hence positive definite. It is easy to show that a sufficient condition for this is that the matrix $\Sigma_W$ has full rank which ensures that the driving noise has full support on $\mathbb{R}^{2d}$ and not just a subspace.

In the construction of a diffusion model, we typically have a forward process which adds noise to a dataset given by \eqref{eqn:SDE}. This forward process is typically chosen in such a way that it is converges to a well defined, easy to sample from distribution, e.g., a normal distribution $\mathcal{N}(0,I_d)$.  We then use this stationary distribution as a prior for the reverse process. We use the same time coordinate $t\in[0,T]$ for both processes, the forward noising process runs from $0\rightarrow T$ and sampling/ denoising process runs back from $T\rightarrow 0$. For an It\^o diffusion $dZ=b(Z,t)\,dt+\Sigma_W\,dW$ with drift term $b(Z,t)$, Anderson's time reversal theorem \cite{Anderson1982-ll} states that there exists a time-reversed process which satisfies
\begin{align}
\label{eqn:ReverseSDE}
d\tilde Z(t)=\Big(b(\tilde Z(t),t)-\sigma\,\nabla_{\tilde Z}\log p_t^{(n)}(\tilde Z(t))\Big)\,dt+\Sigma_W\,d\tilde W(t),
\end{align}
where $p_t^{(n)}$ is the empirical density of $Z(t)$ induced by the empirical initial distribution $p_0^{(n)}$ and $\sigma$ is the diffusion matrix $\sigma=\Sigma_W\Sigma_W^T$ and $dt$ refers to a negative time step. In contrast to the forward process, an additional term which is known as the score function $\nabla_{\tilde{Z}} \log p_t$ arises. In the context of score based diffusion models, the score $\nabla\log p_t^{(n)}$ is approximated by a neural network and has a loss function given by score matching \cite{hyvarinen2005estimation, vincent2011connection, de2021diffusion}. In the analysis of this paper, we assume that one has access to the exact $\nabla\log p_t^{(n)}$ corresponding to $p_0^{(n)}$. In the literature, this is referred to as exact empirical score \cite{cattiaux2023time,haussmann1986time, biroli2024}. 

As previously noted, for the linear SDE \eqref{eqn:SDE}, the transition density is a Gaussian process:
\begin{align}
\label{eqn:PTrans}
K_t(\tilde z\mid z)=\frac{1}{(2\pi)^{d}\sqrt{\det Q(t)}}
\exp\!\left(-\frac{1}{2}(\tilde z-e^{Mt}z)^\intercal Q(t)^{-1}(\tilde z-e^{Mt}z)\right),
\end{align}
where $Z\in\mathbb{R}^{2d}$ and thus the normalization uses $(2\pi)^{d}=(2\pi)^{(2d)/2}$, and $Q(t)\in\mathbb{R}^{2d\times 2d}$ is the transition covariance \eqref{eqn:CondCov}.

Accordingly,
\begin{align}
p_t^{(n)}(\tilde z)=\int K_t(\tilde z\mid z)\,p_0^{(n)}(z)\,dz.
\end{align}
For clarity of analysis, we consider training data that can be separated into different classes. Without loss of generality, we will limit the discussion to two distinct classes however this can be easily generalized to more classes. For the  underlying initial population distribution, we choose a two component Gaussian mixture with equal weights, 
\begin{align}
    P_0(z)=\frac{1}{2}\CN(z;\mu(0), \Sigma(0))+\frac{1}{2}\CN(z;-\mu(0), \Sigma(0)),
\end{align}
where $\mu\in\MR^{2d}$ and 
\begin{align}
\label{eqn:GaussianInit}
    \mu(0)=\begin{pmatrix}
        \mu_x(0)\\
        \mu_y(0)
    \end{pmatrix}, \quad \mu_x, \mu_y \in \MR^d, \quad \Sigma(0)=\begin{pmatrix}
        \sigma_x^2 & 0 \\
        0 & \sigma_y^2 
    \end{pmatrix}\otimes I_d.
\end{align}
The training data consists of $n$ i.i.d. samples $\{z_1,\dots,z_n\}\sim P_0$, hence
\begin{align}
    p^{(n)}_0(z)=\frac{1}{n}\sum_{i=1}^n\delta(z-z_i)
\end{align}
Since the forward process is a Gaussian process \eqref{eqn:PTrans}, each component remains Gaussian with mean drifted by \eqref{eqn:MeanDrift}
\begin{align}
    \mu(t)=e^{Mt}\mu(0),
\end{align}
we define the drifted initial covariance
\begin{align}
    S(t)=e^{Mt}\Sigma(0)e^{M^\intercal t}
\end{align}
while the process accumulates noise given by $Q(t)$ \eqref{eqn:CondCov}. 

The empirical probability distribution becomes a mixture of drifted kernels with width $Q(t)$
\begin{align}
\label{eqn:EmpiricalProb}
    p_t^{(n)}(z)=\frac{1}{n}\sum_{i=1}^n\CN(z; e^{Mt}z_i, Q(t)).
\end{align}
By contrast, the population probability distribution combines the drifted width and the noise, with a diffusion kernel $C(t)=S(t)+Q(t)$,
\begin{align}
\label{eqn:PopulationProb}
    P_t(z)=\frac{1}{2}\CN(z; \mu(t), C(t))+\frac{1}{2}\CN(z; -\mu(t),C(t)).
\end{align}
Depending on the regime of analysis we will either use the population or empirical distribution. In the speciation regime, thresholds are controlled primarily by the separation of the two population components and can therefore be expressed using the parametric form of $P_t$. In contrast, collapse corresponds to an extremal-statistics transition where $p_t^{(n)}$ becomes dominated by a small number of mixture terms. Capturing this requires retaining the discrete empirical sum and the correct kernel width $Q(t)$.

We start with a discussion of the computation of speciation time. For the population mixture $P_t$ in \eqref{eqn:PopulationProb}, completing the square gives
\begin{align}
P_t(z)=\frac{1}{(2\pi)^{d}\sqrt{\det C(t)}}\exp\!\left(-\frac{1}{2}z^\intercal C(t)^{-1}z-\frac{1}{2}\mu(t)^\intercal C(t)^{-1}\mu(t)\right)
\cosh\!\left(\mu(t)^\intercal C(t)^{-1}z\right),
\end{align}
hence
\begin{align}
\label{eqn:LogPop}
\log P_t(z)=\mathrm{const}(t)-\frac{1}{2}z^\intercal C(t)^{-1}z+\log\cosh\!\left(\mu(t)^\intercal C(t)^{-1}z\right),
\end{align}
where $\mathrm{const}(t)$ refers to $z$ independent terms. Substituting \eqref{eqn:LogPop} into the reverse drift in \eqref{eqn:ReverseSDE} (with $a=\sigma_W^2 I$) yields
\begin{align}
\label{eqn:RevDrift}
b_{\mathrm{rev}}(z,t)
&=-Mz+\sigma_W^2\nabla_z\log P_t(z)\nonumber\\
&=-(M+\sigma_W^2C(t)^{-1})z+\sigma_W^2\,C(t)^{-1}\mu(t)\,\tanh\!\big(\mu(t)^\intercal C(t)^{-1}z\big).
\end{align}

If $M$ is symmetric (as in \eqref{eqn:SymMandS}), then $b_{\mathrm{rev}}(z,t)$ is an irrotational vector field. Consequently, the drift  admits a scalar potential $V_t$ which allows us to reformulate the reverse dynamics as a gradient flow $b_{\mathrm{rev}}=-\nabla V_t$.
Up to z independent terms, one convenient choice is
\begin{align}
\label{eqn:SymPotential}
V_t(z)=\frac{1}{2}z^\intercal Mz+\frac{\sigma_W^2}{2}z^\intercal C(t)^{-1}z-\sigma_W^2\log\cosh\!\left(\mu(t)^\intercal C(t)^{-1}z\right).
\end{align}
In the anisotropic case \eqref{eqn:AsymMandS}, the drift is generally non-conservative and no scalar potential exists. Therefore, we define speciation time in a potential independent way as a bifurcation point where the deterministic drift field $b_{\mathrm{rev}}(z,t)$ acquires additional fixed points. In the symmetric case, this bifurcation coincides with the emergence of a double-well structure in $V_t$ as explored in \cite{biroli2024, yu2025nonequilbrium}. Therefore, this critical time corresponds to transition from the regime where trajectories are given by random Brownian motion to a regime where trajectories split into clusters.

The fixed points of the reverse drift satisfy $b_{\mathrm{rev}}(z,t)=0$, i.e.,
\begin{align}
\label{eqn:FixedPointEq}
\big(M+\sigma_W^2C(t)^{-1}\big)z=\sigma_W^2\,C(t)^{-1}\mu(t)\,\tanh\!\big(\mu(t)^\intercal C(t)^{-1}z\big).
\end{align}
We expect that for real data, the distribution is highly concentrated and hence the precision matrix has very large positive eigenvalues initially and hence $\big(M+\sigma_W^2C(t)^{-1}\big)$ can be approximated by a positive definite matrix. As the diffusion time progresses, the OU process is chosen to reach a stationary process such that at any intermediate times, invertibility is ensured. This can enforced numerically provided that the data variance is smaller than the variance of the stationary distribution. We clean this equation up by writing
\begin{align}
A(t):=\big(M+\sigma_W^2C(t)^{-1}\big)^{-1}\sigma_W^2C(t)^{-1}.
\end{align}
Then \eqref{eqn:FixedPointEq} is equivalent to
\begin{align}
z=A(t)\,\mu(t)\,\tanh(u), \quad \text{where} \quad u:=\mu(t)^\intercal C(t)^{-1}z.
\end{align}
Multiplying by $\mu(t)^\intercal C(t)^{-1}$ yields a scalar self-consistency equation
\begin{align}
u=\kappa(t)\,\tanh(u),
\end{align}
with
\begin{align}
\label{eqn:KappaDef}
\kappa(t):=\mu(t)^\intercal C(t)^{-1}A(t)\mu(t).
\end{align}
Besides the trivial solution $u=0$, non-zero solutions appear via a pitchfork bifurcation at $\kappa(t)=1$.
Expanding near $u=0$ gives
\begin{align}
u-\kappa\tanh(u)=u\Big(1-\kappa+\frac{\kappa}{3}u^2+\mathcal{O}(u^4)\Big),
\end{align}
so additional stable fixed points exist if and only if $\kappa(t)>1$. We therefore define the \emph{speciation time} $t_S$ by
\begin{align}
\label{eqn:SpeciationTime}
\kappa(t_S)=1.
\end{align}
Next, we discuss the derivation of the collapse time, which marks the transition to a regime where the diffusion model exhibits memorization. In this regime, the reverse time trajectories flow toward specific training data points, and the model can no longer generalize to unseen data. Mathematically, this corresponds to the point where the empirical distribution diverges from the underlying population distribution.

To derive collapse time, we follow \cite{biroli2024} and we employ techniques from the Random Energy Model (REM) \cite{derrida1981random, mezard1987spin, ruelle1987mathematical, mezard2009information}, which is a disordered statistical physics models known to exhibits phase transitions e.g., it was used for analysis of spin glasses. In this approach, the collapse time corresponds to the condensation transition time when the system freezes to its lowest energy states and the partition function is no longer dominated by the entropy of the population states but instead by the energy of the a few specific training examples. This memorization phenomenon can be clearly seen from empirical probability distribution. Using the diffusion kernel \eqref{eqn:PTrans}, the posterior weights over training indices can be expressed as  
\begin{align}
    w_i(\tilde{z})=\frac{K_t(\tilde{z}|z_i)}{\sum_{j=1}^nK_t(\tilde{z}|z_j)},
\end{align}
where index $i$ labels a training sample $\{z_i\}_{i=1}^n$ and $z_i\in \MR^{2d}$. The model has an ability to generalize when all weights contribute comparably but once it localizes on particular index $i^\star$ i.e., $w_{i^\star}(z)\approx 1$ it memorizes training point. If we denote $\varepsilon_t(\tilde{z},z)=-\frac{1}{2d}\log K_t(\tilde{z}|z)$, then we can write the partition function as
\begin{align}
    \label{eqn:DiffModelPartition}
    Z_t(\tilde{z})=\sum_{i=1}^ne^{-2d\varepsilon_t(\tilde{z},z_i)}.
\end{align}

We will work in the thermodynamic limit where we have asymptotically large d and n with a fixed ratio
\begin{align}
\label{eqn:AlphaDef}
    \alpha=\frac{1}{2d}\log n.
\end{align}
In addition to this, we will assume that $\{\varepsilon_t(\tilde{z},z_i)\}_{i=1}^n$ will behave as independent and identically distributed random variables. We introduce an inverse temperature $\beta$ and we assume there exists a convex, differentiable function
\begin{align}
    \label{eqn:CumulantGen}
    \Lambda_t(\beta)=\lim_{d\rightarrow\infty}\frac{1}{2d}\log\EV[e^{-\beta2d \varepsilon_t}],
\end{align}
which is known as the scaled cumulant generating function of $\epsilon_t$.
By the G\"artner-Ellis theorem, the rate function $I_t$ is then given by the Legendre-Fenchel transform
\begin{align}
\label{eqn:RateFcnDef}
    I_t(u)=\sup_{\beta\in \MR}(-u \beta-\Lambda_t(\beta)),
\end{align}
where the rate function by the large deviation principle is related to a probability distribution of energy levels
\begin{align}
    \mathbb{P}(\varepsilon_t\in [u,u+du])\approx e^{-2dI_t(u)}.
\end{align}
Let $n_t(u)du=\#\{i:\varepsilon_i\in[u,u+du]\}$ denote the density of states. We can use rate function to obtain an expectation value of density of states
\begin{align}
\label{eqn:NumOfStates}
    \EV[n_t(u)]=e^{2d(\alpha-I_t(u))}.
\end{align}
The partition function can be written exactly as
\begin{align}
    Z(\beta)=\int_{\mathbb{R}}du\, n_t(u)e^{-\beta 2du}.
\end{align}
The quenched free energy is then given by
\begin{align}
    \label{eqn:FreeEnergy}\phi_t(\beta)=\lim_{d\rightarrow\infty}\frac{1}{2d}\EV[ \log Z(\beta)]=\sup_{u\in \CD_\varepsilon}(\alpha-I_t(u)-\beta u),
\end{align}
where $\CD_\varepsilon$ is a set of available energy levels and we used the Laplace approximation to replace density of states with typical value \eqref{eqn:NumOfStates}. This set is bounded by a lower limit such that the expectation of number of states becomes of order $1$. Since there $n=e^{2d\alpha}$ samples, from \eqref{eqn:NumOfStates} the lower bound is given by
\begin{align}
\label{eqn:RateLowerBound}
    I_t(u_{\min})=\alpha,
\end{align}
if a supremum $u^\star$ of \eqref{eqn:FreeEnergy}, which is given by
\begin{align}
    \frac{dI_t}{du}(u^\star)=-\beta,
\end{align}
is above $u_{\min}$, then exponentially many states contribute to the partition function. From \eqref{eqn:RateFcnDef}
we find
\begin{align}
    \phi_t(\beta)=\alpha+\Lambda_t(\beta).
\end{align}
However, for larger $\beta$ (lower temperature), the energies will be fixed by a lower bound
\begin{align}
    u^\star&=u_{\min}, \\
    \phi_t(\beta)&=-\beta u_{\min},
\end{align}
this is the frozen phase. The system exhibits phase transition and only a small set of configurations is allowed. The condensation point $\beta_c$ is defined such that
\begin{align}
    I_t(u_{\min})=\alpha, \quad \frac{dI_t}{du} (u_{\min})=-\beta_c,
\end{align}
from Legendre transformation $u_{\min}=-\Lambda'_t(\beta_c)$ where prime denotes differentiation.

Since for the partition function \eqref{eqn:DiffModelPartition}, we have  $\beta=1$, we therefore have for a liquid phase
\begin{align}
    \phi_t=\alpha+\Lambda_t(1),
\end{align}
and for a condensation phase
\begin{align}
    \phi_t=-u_{\min},
\end{align}
the collapse time corresponds to a moment when only a small set of states can be realized so it is given by a solution of $\beta_c(t_C)=1$ which is equivalent to
\begin{align}
\label{eqn:CritCond}
    I_{t_C}(u_{\min})&=\alpha, \\
    u_{\min}&=-\Lambda_{t_C}'(1).
\end{align}
Using the REM model, we transitioned from the computation of collapsed posterior weight to a discussion of energy landscape. At large time, many terms contribute and weights are spread. As we approach $t=0$ the lowest energy dominate the sum and the model collapse to best matching training point. We now proceed to the discussion of particular choices of the relaxation matrix $M$.

\section{The Symmetric Relaxation Matrix}
\label{sec:IsoCoup}
We now focus on the case of a symmetric relaxation matrix $M$. This choice will allow us to diagonalize the coupled system into orthogonal eigenmodes which offer a clear physical interpretation of the generation process. To visualize this spectral decomposition, consider the joint generation of text and images. Rather than viewing the processes as being separate, we can decompose their generating dynamics into common modes representing the shared semantic content/ alignment between the caption and the visual scene and a difference mode that represents the discrepancies between the modalities.

Explicitly, we can diagonalize the  \eqref{eqn:SymMandS} with the choice of normalized eigenvectors
\begin{align}
\label{eqn:PMbasis}
    v_+=\frac{1}{\sqrt{2}}\begin{pmatrix}
        1 \\ 1
    \end{pmatrix}, \qquad
    v_-=\frac{1}{\sqrt{2}}\begin{pmatrix}
        1 \\ -1
    \end{pmatrix},
\end{align}
and their respective eigenvalues 
\begin{align}
    \lambda_+=-\beta+g, \; \text{and},\; \lambda_-=-\beta-g.
\end{align}
We call $v_+=\tfrac{1}{\sqrt{2}}(X+Y)$ a ``common'' mode and $v_-=\tfrac{1}{\sqrt{2}}(X-Y)$ a ``difference'' mode, which we will see have different decay rates. From the construction, we interpret common mode, denoted with $+$ as a one that controls shared content, whereas a difference mode, denoted with $-$, controls disagreement between modalities.  Since we require $\beta>|g|$ for stability, the common mode has smaller decay rate and thus retains signal longer under forward noising process in comparison to the difference mode. Since the end of the generation process is at $t\rightarrow 0$, this implies that the diffusion model generates a hierarchical emergence of features i.e. it determines the shared semantic content first, establishing a consistent global structure before resolving the independent, modality specific details via the difference mode. This separation allows us to analyze the speciation and collapse times for each mode independently and study the corresponding synchronization gaps between the eigenmodes.

Before we proceed, we will derive a formula for general speciation and collapse times in the case of symmetric $M$. We start with a computation of $\kappa(t)=\mu(t)^\intercal C(t)^{-1}\mu(t)$ using the diagonalizing basis. Let us define projection operators from the spectral decomposition \eqref{eqn:PMbasis}
\begin{align}
\label{eqn:ProjOps}
    P_+=v_+v_+^\intercal\otimes I_d, \quad P_-=v_-v_-^\intercal\otimes I_d,
\end{align}
and decompose the initial Gaussian mean $\mu(0)$ as
\begin{align}
    \mu_+(0)=v_+^\intercal \mu(0)=\frac{1}{\sqrt{2}}\left(\mu_x(0)+\mu_y(0)\right), \quad \mu_-(0)=v_-^\intercal \mu(0)=\frac{1}{\sqrt{2}}\left(\mu_x(0)-\mu_y(0)\right),
\end{align}
then the drift equation \eqref{eqn:MeanDrift}
\begin{align}
\label{eqn:ProjMu}
    \mu(t)=e^{Mt}\mu(0)=\left(e^{\lambda_+t}P_++e^{\lambda_-t}P_-\right)\mu(0)=e^{\lambda_+t}\mu_+(0)\otimes v_++e^{\lambda_-t}\mu_-(0)\otimes v_-.
\end{align}
Similarly, we find the eigenvalues of the diffusion kernel
\begin{align}
    c_{\pm}(t)=\sigma^2e^{2\lambda_\pm t}+\sigma_W^2\frac{e^{2\lambda_\pm t}-1}{2\lambda_{\pm}},
\end{align}
where we also assume that $\sigma_x=\sigma_y=\sigma$ in \eqref{eqn:GaussianInit}, so that we can decompose it with projection operators \eqref{eqn:ProjOps}
\begin{align}
    C(t)=c_+(t)P_++c_-(t)P_-,
\end{align}
and find the inverse
\begin{align}
    C(t)^{-1}=\frac{1}{c_+(t)}P_++\frac{1}{c_-(t)}P_-.
\end{align}
In the $\pm$ basis,
\begin{align}
\big(M+\sigma_W^2C(t)^{-1}\big)^{-1}
=
\frac{1}{\lambda_+ + \sigma_W^2/c_+(t)}P_+
+
\frac{1}{\lambda_- + \sigma_W^2/c_-(t)}P_-.
\end{align}
Hence invertibility requires $\lambda_\pm + \sigma_W^2/c_\pm(t)\neq 0$. Moreover, in the symmetric case, the reverse potential \eqref{eqn:SymPotential} is confining at large $\|z\|$ when
\begin{align}
\label{eqn:TailStability}
\lambda_\pm + \frac{\sigma_W^2}{c_\pm(t)} > 0,
\end{align}
A sufficient condition ensuring \eqref{eqn:TailStability} for all $t\in[0,T]$ is
\begin{align}
\label{eqn:InvCond}
\sigma^2 < \frac{\sigma_W^2}{\beta+|g|},
\end{align}
since $c_\pm(t)\ge \sigma^2$ and $\sigma_W^2/c_\pm(t)$ is non-increasing in $t$. Violating \eqref{eqn:TailStability} leads to a non-confining reverse potential tail which we call unstable symmetry breaking and can cause sampling trajectories to diverge \cite{yu2025nonequilbrium}.

Using the orthogonality of the eigenvectors, we obtain for $\kappa$ \eqref{eqn:KappaDef}
\begin{align}
    \kappa(t)=\sigma^2_W\left[\frac{e^{2\lambda_+ t}||\mu_+(0)||^2}{c_+(t)(\lambda_+c_+(t)+\sigma_W^2)}+\frac{e^{2\lambda_- t}||\mu_-(0)||^2}{c_-(t)(\lambda_-c_-(t)+\sigma_W^2)}\right],
\end{align}
where we used \eqref{eqn:ProjMu}. To analyze the system in the high dimension limit as $d\rightarrow \infty$, we introduce a per dimension norm $m_\pm^2=\tfrac{1}{d}||\mu_\pm(0)||^2$ and we rescale $\kappa(t)$ accordingly by a factor of $1/d$ to ensure it remains of order $\mathcal{O}(1)$. Then, we can rewrite the last equation as

\begin{align}
\label{eqn:KappaGeneral}
\kappa(t)=\sigma^2_W\left[\frac{e^{2\lambda_+ t}m_+^2}{c_+(t)(\lambda_+c_+(t)+\sigma_W^2)}+\frac{e^{2\lambda_- t}m_-^2}{c_-(t)(\lambda_-c_-(t)+\sigma_W^2)}\right].
\end{align}

\begin{figure}
    \centering
    \includegraphics[width=0.85\linewidth]{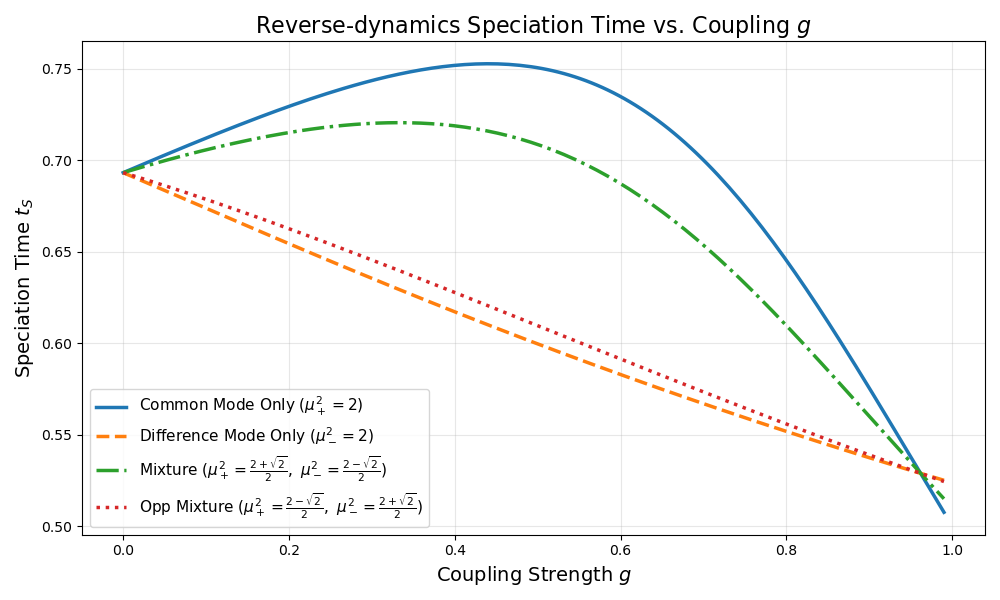}
    \caption{Impact of coupling strength $g$ on speciation time $t_s$. Analytic solutions (solid and dashed) show that the Common Mode only case exhibits a non-monotonic relationship, peaking near $g=0.45$, while the Difference Mode leads to a monotonic decrease in speciation time. Numerical results for mixtures (dotted/dash-dotted) demonstrate how intermediate modal configurations interpolate between these extremes. Parameters are set to $\beta = 1, \sigma_W^2 = 2,$ and $\sigma^2 = 1$.}
    \label{fig:SpeciationTime}
\end{figure}

We can interpret this equation as a sum of signal to noise ratios for the common and difference modes
\begin{align}
\label{eqn:SpecSumSNRs}
    \kappa(t)=\sigma_W^2(\text{SNR}_++\text{SNR}_-).
\end{align}
The equation above reveals that the total signal driving the speciation is additive. However, due to the exponential separation of timescales, the contribution from the common mode dominates the early dynamics. 
If $\kappa(t)<1$ for all $t$, the model loses the ability to distinguish between different classes and the system is stuck in the noise regime. In fact when \eqref{eqn:InvCond} is satisfied then both $\text{SNR}_\pm$ are strictly decreasing functions in time, therefore $\kappa(t)$ achieves its maximum at $t=0$. Therefore, the system will never leave the noise regime if 
\begin{align}
    \kappa(0)=\sigma^2_W\left[\frac{m_+^2}{\sigma^2(\lambda_+\sigma^2+\sigma_W^2)}+\frac{m_-^2}{\sigma^2(\lambda_-\sigma^2+\sigma_W^2)}\right]<1.
\end{align}
In the general case, we cannot solve \eqref{eqn:KappaGeneral} analytically for speciation time. Hence, we first analyze the two opposite cases when either the initial common or difference mode vanishes. For both, we can write analytic expression for speciation time $t_S$ \eqref{eqn:SpeciationTime}
\begin{align}
    1=\sigma_W^2\text{SNR}_{\pm}=\sigma_W^2\frac{e^{2\lambda_\pm t}m_\pm^2}{c_\pm(t)(\lambda_\pm c_\pm(t)+\sigma_W^2)},
\end{align}
which gives a quadratic equation for $e^{2\lambda_\pm t}$, taking the positive root we get
\begin{align}
\label{eqn:AnalyticForm}
    t=\frac{1}{2\lambda_{\pm}}\log\left(\frac{\sigma^2_W\left(m_\pm^2+\sqrt{m_\pm^4+B^2}\right)}{2\lambda_\pm B^2}\right),
\end{align}
where $B=\sigma^2+\frac{\sigma^2_W}{2\lambda_\pm}$. We use this analytic form to plot speciation time $t_S$ versus coupling strength $g$ on \figref{fig:SpeciationTime} for common and difference mode cases. We also include a numerical solution for the mixtures mode cases. 

Clearly, the value of coupling strength $g$ changes the speciation time $t_S$ and this change has a different characteristic for the common and difference mode. In the case of the analytic formula \eqref{eqn:AnalyticForm}, one can find an optimal value of coupling strength $g$ such that speciation time is maximal which corresponds to the earliest time in the generation process. The coupling induces a temporal splitting of the speciation transition. As expected, the common mode whose decay rate is slower has sooner speciation time than a difference mode, however the synchronization gap between these two shrinks as coupling strength grows to the point when strong coupling causes difference mode to speciates before the common one. From a generative standpoint, this identifies a tunable synchronization regime. Coupling can either widen or shrink the time interval in which the global structure is decided while modality specific discrepancies are still unresolved. 

The mixture modes confirm that speciation time $t_S$ is not determined by a single dominant mode unless the initialization is strongly aligned with one eigenspace. For diffusion modeling, this supports the view that coupling acts like a spectral filter. The reverse process selects which collective coordinates become informative first, which organizes the generation based on how the signal projects onto the eigenspaces of the coupled dynamics. 

In addition, the bound \eqref{eqn:InvCond} on the decay rate for the model to transition from the noise regime provides a criterion for when a coupled diffusion model may become numerically unstable or explode in sampling, and it directly links that instability to kernel growth and coupling strength.

In the discussion above, we only focused on nonnegative $g$ since we are interested in the case where both modalities are aligned due to a coupling. In the case of negative $g$ the role of the modes reverses and the difference mode dominates early dynamics. 

Now, we proceed with a discussion of the collapse time i.e. the transition time from regime II to regime III where the diffusion model loses the ability to generalize and instead starts a memorization process. We start with a calculation of cumulant generating function \eqref{eqn:CumulantGen}, where we leave the details of calculation for \appref{app:DerCGF}. Denoting
\begin{align}
\label{eqn:Chi}
    \chi_{\pm}(t)=\frac{\sigma^2}{\sigma_W^2}\frac{2\lambda_{\pm}}{1-e^{-2\lambda_{\pm}t}},
\end{align}
we find
\begin{align}
\label{eqn:LambdaSym}
    \Lambda_t(\beta)=-\frac{1}{4}\log(1+\beta\chi_+)-\frac{1}{4}\log(1+\beta\chi_-)-\frac{1}{4}\frac{\beta(1+\chi_+)}{1+\beta\chi_+}-\frac{1}{4}\frac{\beta(1+\chi_-)}{1+\beta\chi_-},
\end{align}
and the saddle point \eqref{eqn:CritCond} of this particular form of $\Lambda_t(\beta)$ is given by
\begin{align}
    -\Lambda'_t(1)=\frac{1}{2}.
\end{align}
Therefore, the collapse time can be found by solving
\begin{align}
    \alpha=I_{t_C}\left(\frac{1}{2}\right)=-\Lambda_{t_C}(1)-\frac{1}{2},
\end{align}
explicitly we get
\begin{align}
    n^{2/d}=\left(1+\frac{\sigma^2}{\sigma_W^2}\frac{2\lambda_+}{1-e^{-2\lambda_+t_C}}\right)\left(1+\frac{\sigma^2}{\sigma_W^2}\frac{2\lambda_-}{1-e^{-2\lambda_-t_C}}\right),
\end{align}
or if we use decay rate $\tau$
\begin{align}
\label{eqn:ColTimeEqn}
    n^{2/d}=\left(1+\frac{\sigma^2}{\sigma_W^2}\frac{\tau_+}{e^{\tau_+t_C}-1}\right)\left(1+\frac{\sigma^2}{\sigma_W^2}\frac{\tau_-}{e^{\tau_-t_C}-1}\right),
\end{align}
again this is a transcendental equation that for general $g$ needs to be solved numerically. Unlike speciation, which depends on the between-class separation encoded in $\mu$, condensation is controlled by the transition variance $Q(t)$ and the number of training points. Using the diffusion kernel $C(t)$, transition covariance $Q(t)$ and \eqref{eqn:AlphaDef} we can show that \eqref{eqn:ColTimeEqn} is equivalent to
\begin{align}
\label{eqn:CollapseKernel}
    \alpha=\frac{1}{4}\log\left(\frac{C(t_C)}{Q(t_C)}\right)
\end{align}

Notice that, provided the system is in the stable regime, $\chi_{\pm}$ is a strictly decreasing function in $t_C$ and $\tau$ . In fact
\begin{align}
    \lim_{t_C\rightarrow 0^+}\chi_{\pm}=\infty, \; \text{and} \; \lim_{t_C\rightarrow \infty}\chi_{\pm}=0.
\end{align}
Here, we clearly see the curse of dimensionality \cite{donoho2000high, biroli2024} for diffusion models. If $n^{2/d}=1$, there is no finite collapse time and the model cannot generalize at all. For $n^{2/d}>1$, there is an unique collapse time. If we use the inequality $e^x-1\geq x$, we can construct an upper bound for the collapse time
\begin{align}
    t_C\leq t_{\max}=\frac{\sigma^2/\sigma_W^2}{n^{1/d}-1},
\end{align}
this bound is saturated in the case when coupling $g$ vanishes. 

For a symmetric coupling, in a diagonal basis \eqref{eqn:PMbasis}, we could factorize a transition kernel \eqref{eqn:PTrans}
\begin{align}
    K_t(\tilde{z}|z)=K_{+,t}(\tilde{z}_+|z_+)K_{-,t}(\tilde{z}_-|z_-),
\end{align}
where $z_{\pm}$ are constructed using projection operators \eqref{eqn:ProjOps}. Therefore, we can can construct partition functions for $\pm$ modes separately and introduce their respective collapse times. Formally, this transition corresponds to the moment when the posterior weights $w_{i,\pm}(z_\pm)$ for a particular training point $i$ become of order $\mathcal{O}(1)$. In such case, we find $t_{C_\pm}$ by solving
\begin{align}
    n^{1/d}&=1+\frac{\sigma^2}{\sigma_W^2}\frac{\tau_{\pm}}{e^{\tau_\pm t_{C_\pm}}-1}, 
\end{align}
which can be analytically solved to give
\begin{align}
    t_{C_\pm}=\frac{1}{\tau_{\pm}}\log\left(1+\frac{\sigma^2}{\sigma^2_W}\frac{\tau_\pm}{n^{1/d}-1}\right).
\end{align}
\begin{figure}[ht]
    \centering
    \includegraphics[width=0.85\linewidth]{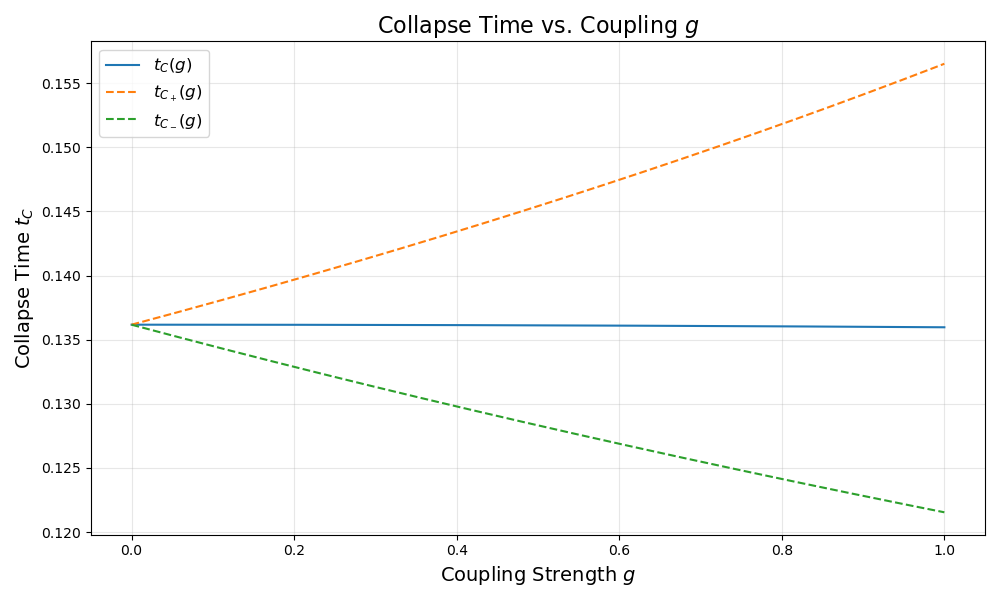}
    \caption{Collapse time vs coupling strength  $g$. We plot analytic solutions for the collapse time of common and difference modes and numerical solutions. For plotting and numerical evaluation, we set $\alpha=1$ and $\sigma^2/\sigma_W^2=1$.}
    \label{fig:CollapseTime}
\end{figure}

In \figref{fig:CollapseTime}, we plot the collapse times for the common mode $t_{C_+}$, the difference mode $t_{C_-}$ and the collapse time of  the joint system as a function of the coupling strength $g$. The graph shows a temporal splitting of the timescales that explains the hierarchical behavior of coupled generative models. Interestingly, $t_C$ remains nearly constant despite an increase of $g$. This stability arises because the contributions $\chi_\pm$ effectively compensate for one another. As a practical guide, this shows that the coupling strength $g$ serves as a robust control parameter which allows us to tune the generative hierarchy without risking a collapse of the entire joint system.

Analogous to the speciation phase, there is a corresponding hierarchy in the memorization process. As the coupling strength $g$ increases, we observe widening of the temporal gap between the modes. The collapse time of a common mode increases implying that the alignment between modalities enters the memorization regime earlier in the generative process. In contrast, the difference mode collapse time decreases, indicating that the fine-grained, modality-specific details remain in the generalization regime until late times in the generative process. This separation creates a temporal window $t_{C_-}<t<t_{C_+}$ where the model exhibits hybrid behavior, i.e., the global structure is now memorized but there are still local details that are being memorized. This analysis suggests that the coupling strength $g$ additionally acts as a control parameter for the generalization bandwidth.

\section{The Anisotropic Relaxation Matrix}
\label{sec:AsymCoup}
We now proceed with a discussion of the anisotropic relaxation matrix case defined in \eqref{eqn:AsymMandS}. This setup is particularly relevant for conditional generative tasks, such as text-to-image generation, where a fixed context modality $X$ guides the generation of a target modality $Y$. In this model, there is a unidirectional information flow from $X$ to $Y$ without reciprocal feedback via the coupling $g$. 

We start with the computation of the speciation time $t_S$ using the condition $\kappa(t_S)=1$ \eqref{eqn:SpeciationTime}. For notational convenience, we define the operator
\begin{align}
\label{eqn:MatrixK}
    K(t)=C(t)^{-1}\left(M+\sigma^2_W C(t)\right)^{-1}C(t)^{-1},
\end{align}
which is generally non-symmetric and where $M$ is given by \eqref{eqn:AsymMandS} and the explicit form of the matrix elements of the diffusion kernel $C(t)$ is provided in \appref{app:ExpCt}. In the appendix, we also provide explicit formulas for elements of $K(t)$.  Solving the coupled drift dynamics yields the following explicit mean trajectories
\begin{align}
    \mu_x(t)&=e^{-\beta t}\mu_x(0), \\
    \mu_y(t)&=e^{-\beta t}(\mu_y(0)+gt\mu_x(0)).
\end{align}
Using the per-dimension norms $m_{x,y}^2=\frac{1}{d}||\mu_{x,y}(0)||^2$ and the alignment angle $\theta$ defined by
\begin{align}
    \label{eqn:ThetaAngle}
    \cos\theta=\frac{\mu_x(0)\cdot\mu_y(0)}{||\mu_x(0)||\,||\mu_y(0)||},
\end{align}
we can solve the speciation condition numerically. 

\begin{figure}[ht]
    \centering
    \includegraphics[width=0.85\linewidth]{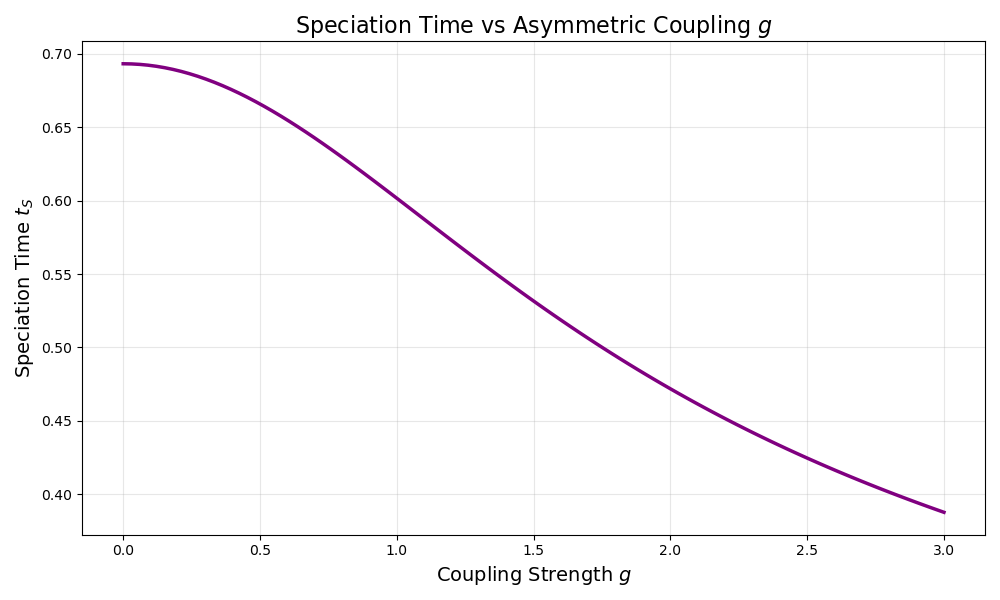}
    \caption{Numerical solution for speciation time vs. coupling strength $g$ in the uninformative context regime ($\mu_x(0)=0$ and $m_y^2=2$). The monotonic decrease in $t_S$ indicates that stronger coupling delays class speciation toward the end of the reverse process (closer to $t=0$). }
    \label{fig:SpecTimeAsym}
\end{figure}

We analyze two distinct regimes. First, we consider an uninformative context scenario where the conditioning signal $X$ has negligible between-class separation where $\mu_x(0)=0$, while the target $Y$ carries the label information $m_y^2 > 0$. This effectively models cases where the context provides general structure but cannot be used to distinguish between different class identities. The results are plotted in \figref{fig:SpecTimeAsym} for the case when $\sigma_W^2=2$, $\sigma_x^2=\sigma_y^2=1$ and $\beta=1$.
As shown in \figref{fig:SpecTimeAsym}, the speciation time $t_S$ decreases monotonically with the coupling strength $g$. Since sampling proceeds from $T\to 0$, smaller $t_S$ means that speciation occurs later in the reverse process. In this uninformative context regime $\mu_x(0)=0$, the mean trajectory $\mu_y(t)=e^{-\beta t}\mu_y(0)$ is independent of $g$. Nevertheless, the coupling modifies the reverse time bifurcation through the covariance,  $g$ increases the effective uncertainty seen by the reverse drift by first mixing $X$ variance into the $Y$ channel through the drifted initial covariance $S(t)=e^{Mt}\Sigma(0)e^{M^\intercal t}$, and secondly, inducing correlated noise through the transition covariance $Q(t)$. Both effects inflate the variance of $Y$ and thereby reduce $\kappa(t)$, delaying  or completely preventing the onset of additional fixed points. In this regime, increasing $g$ acts primarily as a noise drag.

\begin{figure}[ht]
    \centering
    \includegraphics[width=0.85\linewidth]{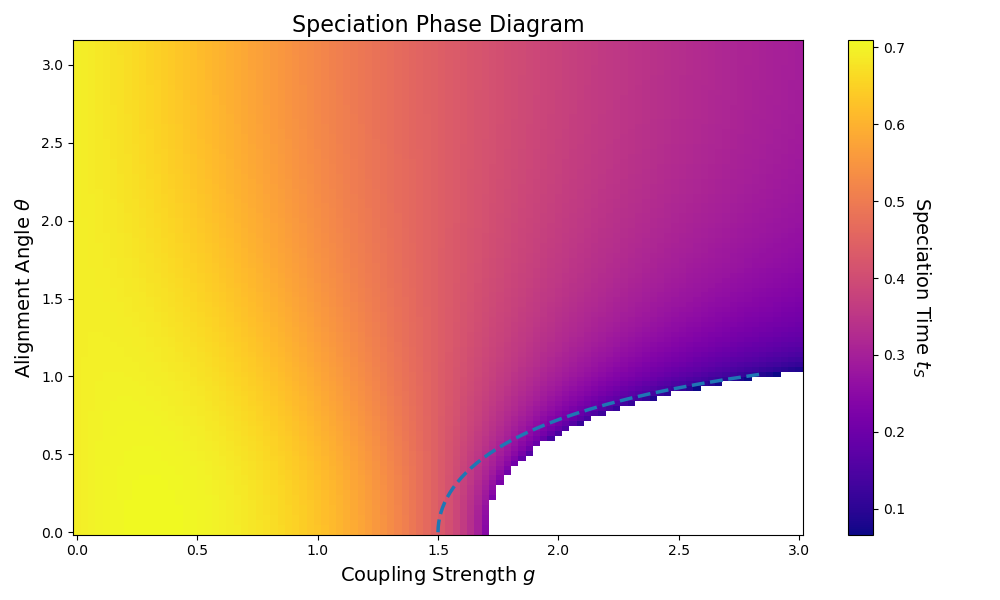}
    \caption{Numerical solution for speciation time vs. coupling constant $g$ and angle $\theta$ in the case of $m_x^2=m_y^2=1$. For plotting we set $\sigma_W^2=2$, $\sigma_x^2=\sigma_y^2=1$ and $\beta=1$. White color labels region where $\kappa(t)<1$ for all $t$. The blue dashed line indicates an approximate $g_{\text{crit}}(\theta)$.}
    \label{fig:PhaseDiag}
\end{figure}

Next, we consider the general case where both modalities carry information with varying alignment angle $\theta$. For simplicity, we consider the case where they both have the same norm $m^2_x=m^2_y=1$. The phase diagram of speciation time $t_S(g, \theta)$ is presented in \figref{fig:PhaseDiag}. White color indicates the no-speciation regime
\begin{align}
\label{eqn:nospec_def}
\sup_{t\in[0,T]}\kappa(t)\le 1,
\end{align}
in which the reverse drift field does not acquire additional fixed points and the dynamics remains effectively unimodal (i.e. noise-dominated) throughout the sampling.
Unlike the symmetric case, there is no additional regime where the coupling accelerates speciation by increasing $t_S$. Instead, increasing $g$ generally delays speciation. This effect is strongly dependent on alignment. When the means are aligned $\theta\approx 0$, the system is initially less sensitive to coupling. However, around $g\approx 1.0$ $t_S$ drops rapidly from $\approx 0.6$  to $\approx 0.1$ and eventually enters an no-speciation regime for $g \gtrsim 1.7$. When the means are anti-aligned $\theta\approx \pi$, the initial speciation time is lower due to signal interference. However, the sensitivity to $g$ is reduced, the speciation time decreases more gradually, and the system avoids failure to bifurcate within the plotted range.

In the symmetric case $\sigma_x^2=\sigma_y^2=\sigma^2$, which we use for our experiments, one has $C(0)=\sigma^2 I$ and hence
\begin{align}
A(0)=\big(M+rI\big)^{-1}\,rI,\qquad r:=\sigma_W^2/\sigma^2.
\end{align}
\begin{align}
\label{eqn:kappa0_asym}
\kappa(0)=\frac{r}{\sigma^2(r-\beta)}\big(m_x^2+m_y^2\big)
-\frac{rg}{\sigma^2(r-\beta)^2}\,m_x m_y\cos\theta,
\end{align}
where $m_{x,y}^2=\frac{1}{d}\|\mu_{x,y}(0)\|^2$.
Equation \eqref{eqn:kappa0_asym} makes the alignment sensitivity explicit. For the case of aligned means where$\cos\theta>0$ , increasing $g$ decreases $\kappa(0)$, while for anti aligned means where $\cos\theta<0$ , it increases $\kappa(0)$.
In practice we observe that $\kappa(t)$ is maximized near $t=0$ across the parameter ranges of \figref{fig:PhaseDiag}, so $\kappa(0)\le 1$ provides an accurate diagnostic for the no-speciation regime \eqref{eqn:nospec_def}, and yields an explicit approximate phase boundary $g_{\mathrm{crit}}(\theta)$ via $\kappa(0)=1$, which on \figref{fig:PhaseDiag} is marked with blue dashed line and for larger values of $\theta$ correctly approximate the boundary of the white region.

The phase diagram suggests that large effective coupling can suppress bifurcation when the conditional and unconditional directions are locally aligned, pushing the system into the no-speciation regime \eqref{eqn:nospec_def}.
This motivates treating the coupling strength as a sampling-time control, analogous to guidance scaling as for example in Classifier-Free Guidance (CFG) scales \cite{ho2022classifier, chung2024cfg}. In \appref{sec:toy_ou_experiment} we discuss an exact score numerical experiment in which we examine different scheduling patterns for coupling and tests their results with respect to $\theta$. We show that moderate coupling provides better results in the misaligned phase ($\theta>\pi/2$).   

We now move on to the discussion of transition between regime II and regime III. In the asymmetric case, we are interested in two different collapse times. First we consider a joint collapse time as given by \eqref{eqn:CollapseKernel} 
\begin{align}
    \alpha=\frac{1}{4}\log\left(\frac{\det C(t_C)}{\det Q(t_C)}\right)
\end{align}
but we can also consider conditional collapse time
\begin{align}
    \alpha_C=\frac{1}{2}\log\left(\frac{C_{y|x}(t_{C,y|x})}{Q_{y|x}(t_{C,y|x})}\right),
\end{align}
where the prefactor is twice since effective dimension is $d$ and we compute conditional covariances using the Schur complement
\begin{align}
    C_{y|x}(t)&=C_{22}(t)-\frac{C_{12}(t)^2}{C_{11}(t)} \\
    Q_{y|x}(t)&=q_{22}(t)-\frac{q_{12}(t)^2}{q_{11}(t)}.
\end{align}
\begin{figure}[ht]
    \centering
    \includegraphics[width=0.85\linewidth]{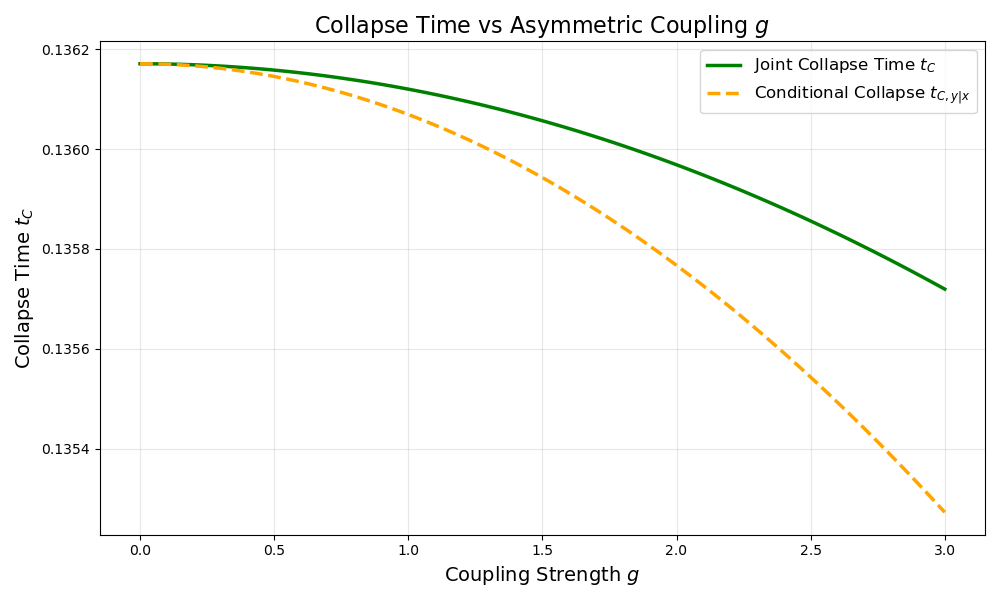}
    \caption{Numerical solutions for joint collapse time $t_C$ and conditional collapse time $t_{C,y|x}$. For plotting we set $\sigma_W^2=1$, $\sigma_x^2=\sigma_y^2=1$, $\beta=1$ and $\alpha=1$.}
    \label{fig:AsymCollapse}
\end{figure}
We show the behavior of the joint collapse time $t_C$ and conditional collapse time $t_{C,y|x}$ on \figref{fig:AsymCollapse}. As in the case of symmetric coupling, the condensation time depends weakly on $g$. The conditional collapse time $t_{C,y|x}$ is smaller than the joint $t_C$, reflecting that conditioning reduces the effective uncertainty in $Y$.  

\section{MNIST Synchronization Experiment}
\label{sec:MinstExp}
Our analysis of the symmetric relaxation matrix in \secref{sec:IsoCoup} reveals that the generation process undergoes two distinct transitions, i.e., speciation and collapse which occur at two distinct time scales. In the case of speciation, the common mode stabilizes first and is followed by the difference mode. Here, the stabilization meaning the transition from the high noise regime I into the clustered structure of regime II. This implies the existence of a temporal window which we refer to as a synchronization gap. When we introduced common and difference modes, we discussed that the information can be understood in a hierarchical fashion where the global structure is determined before resolving the specific, higher frequency signals between modalities i.e. the fine details of the generation. In this case, the synchronization gap can be understood as a period where the global structure is already decided while modality specific discrepancies remain unstable. In generative modeling, such a gap is expected to manifest as a transient misalignment and, crucially, as a selective susceptibility of the difference mode to perturbations during that window.

To validate our theoretical framework, we construct a minimal image experiment to investigate the predictions in the case of the symmetric relaxation matrix.  In order to do this, we do not directly sample the theoretical Ornstein-Uhlenbeck SDE but instead we will implement a standard diffusion model architecture whose data we can split into a two channel state. Our objective in this experiment is not about sample quality but instead about seeing the temporal ordering between modes.

Let $x \in [0,1]^{28\times 28}$ be an MNIST image \cite{lecun2002gradient}. We define a deterministic, pixel aligned modality transform $T$ selected from a set of operations including edge mapping, inversion, morphological dilation or blur. From $x$, we form a two channel state
\begin{align}
z_0 = \begin{bmatrix} x_A \\ x_B \end{bmatrix} \in \mathbb{R}^{2\times 28\times 28}.
\end{align}

We introduce two different sample sets. First we introduce \textit{MAIN pairing}. For each underlying digit image $x$, we set $(x_A,x_B)=(x, T(x))$ such that they are both in the same label class and then randomly swap the channels with uniform probability. This enforces an explicit $A/B$ exchange symmetry at the dataset level and prevents the model from relying on a fixed channel identity.

Our second pairing will be called \textit{CTRL pairing} where we break content alignment while preserving per channel marginals by sampling $x_A$ and $x_B$ from different MNIST indices with $x_A=x_i$ and $x_B=T(x_{\pi(i)})$, where $\pi$ is a random permutation refreshed each epoch constrained such that $y_i \neq y_{\pi(i)}$ ensuring disjoint classes. We apply the same random $A/B$ random swap as in the MAIN pairing. Consequently, any observed mode ordering cannot be attributed to fixed channel conventions or marginal statistics but strictly to the presence or absence of cross-channel semantic alignment.

As in \secref{sec:IsoCoup}, we diagonalize the basis using the change of variables
\begin{align}
u = \frac{x_A + x_B}{\sqrt{2}},\qquad
v = \frac{x_A - x_B}{\sqrt{2}}.
\end{align}
Using the new basis, we denote stacked representation as $z^{(uv)} = [u, v]^\top$. By construction,  the common mode $u$ captures structure/content shared across channels while difference mode $v$ isolates misalignment between channels.

In the diagonal $(u,v)$ basis, where components decouple, the scalar coupling parameter $g\in[0,1)$ is implemented by modifying the diffusion noise covariance. Specifically, at each diffusion step, we draw
\begin{align}
\label{eqn:NoiseLaw}
\varepsilon_u \sim \mathcal{N}(0,(1-g)\,I),\qquad
\varepsilon_v \sim \mathcal{N}(0,(1+g)\,I),
\end{align}
independently over pixels, and then map $(u,v)$ back to $(x_A,x_B)$ coordinates by the inverse change of variables.

Equivalently, in $(x_A,x_B)$ coordinates, the noise has unit per-channel variance but a tunable cross channel correlation
\begin{align}
\label{eqn:EpsUV}
\mathrm{Var}(\varepsilon_A)=\mathrm{Var}(\varepsilon_B)=1,\qquad
\mathrm{Cov}(\varepsilon_A,\varepsilon_B)=-g.
\end{align}
We employ a standard variance preserving discrete diffusion with $T$ timesteps and linear $\beta$ schedule. Defining $\alpha_t=1-\beta_t$ and the cumulative product $\bar{\alpha}_t=\prod_{s=0}^t \alpha_s$. The forward noising process is given by
\begin{align}
z_t = \sqrt{\bar{\alpha}_t}\,z_0 + \sqrt{1-\bar{\alpha}_t}\varepsilon,
\end{align}
where $\varepsilon$ is sampled with the anisotropic variance described above. When the coupling strength $g=0$, this setup reduces down to standard isotropic Gaussian noise in both modes. Under this variance preserving formulation, the per mode noise variance for each mode at a time step $t$ scales as $\mathrm{Var}(z^{(m)}_t| z^{(m)}_0)\propto (1-\bar\alpha_t)\sigma_m^2$, with $\sigma_u^2=1-g$ and $\sigma_v^2=1+g$.
Consequently, a convenient proxy for per-mode SNR therefore satisfies
\begin{equation}
\frac{\mathrm{SNR}_u(t)}{\mathrm{SNR}_v(t)}\propto \frac{1+g}{1-g},
\end{equation}
predicting that increasing $g$ monotonically widens the ordering i.e. $u$ stabilizes earlier than $v$. This is the minimal mechanism needed to test different speciation times for both modes and the synchronization gap hypothesis.

We train an unconditional $\varepsilon$-prediction diffusion model on $z_t\in\mathbb{R}^{2\times 28\times 28}$.
The architecture is a compact U-Net  \cite{ronneberger2015u} with sinusoidal time embedding. It takes $(z_t,t)$ and outputs
$\hat{\varepsilon}_\theta(z_t,t)\in\mathbb{R}^{2\times 28\times 28}$. The training minimizes the standard mean-squared error
\begin{align}
\mathcal{L}(\theta)
=
\mathbb{E}_{z_0,\,t,\,\varepsilon}
\Big[\big\|\hat{\varepsilon}_\theta(z_t,t) - \varepsilon\big\|_2^2\Big],
\end{align}
with $t$ sampled uniformly from $\{0,\dots,T-1\}$ and $\varepsilon$ drawn from the mode-shaped noise law \eqref{eqn:NoiseLaw}. We keep an exponential moving average of parameters for evaluation to reduce sampling variance. We note that because of our target $\epsilon$ is anisotropic in the variance, the squared magnitude of the errors will be inherently larger for the $v$ modes since the target values are typically larger implying that the corresponding gradient will be dominated by errors in the $v$ mode. 

\subsection{Protocol I: Deterministic Synchronization Diagnostics}

To analyze mode resolved stabilization along the reverse process, we employ a DDIM deterministic sampling \cite{song2020denoising},
initialized from $z_{T-1}$ drawn from the same mode shaped noise distribution.
At each reverse step $t$, we compute the predicted clean state in $(u,v)$ coordinates:
\begin{align}
\widehat{z}_0^{(uv)}(t)
=
\frac{z_t^{(uv)} - \sqrt{1-\bar{\alpha}_t}\,\hat{\varepsilon}^{(uv)}_\theta(z_t,t)}{\sqrt{\bar{\alpha}_t}}.
\end{align}
We log $(x_A,x_B)$ along the full time reversed trajectory with a chosen timestep and then project the logged predictions into $u(t)$ and $v(t)$ for mode resolved analysis. Let $m(t)\in\{u(t),v(t)\}$ denote a logged mode prediction at reverse time $t$. We quantify stabilization and observe the synchronization gap in two complementary ways. Let $m_\star=m(t_{\mathrm{final}})$ be the last logged prediction near $t=0$. Flattening images into vectors, we may use the cosine similarity
\begin{align}
\label{eqn:CosSim}
\mathrm{Cos}_m(t)
=
\mathbb{E}\Big[\cos\big(m(t),m_\star\big)\Big].
\end{align}
This is metric is invariant to signal magnitude and consequently any value approaching 1 indicates that the mode vector has been aligned with the ground truth. For a threshold $\tau\in\{0.90,0.95,0.98\}$, we define the crossing time as the earliest reverse step at which alignment exceeds $\tau$
\begin{equation}
\label{eqn:TimeGaps}
t_u(\tau):=\max\{t:\text{Cos}_u(t)\ge \tau\},\qquad
t_v(\tau):=\max\{t:\text{Cos}_v(t)\ge \tau\}.
\end{equation}
We can then find a synchronization gap at $\tau$
\begin{equation}
\Delta t(\tau) := t_v(\tau)-t_u(\tau).
\end{equation}
Thus $\Delta t(\tau)<0$ means the common mode $u$ stabilizes earlier than the difference mode $v$ by $|\Delta t|$ reverse steps.

To provide an interpretable counterpart to our metrics, we visualize for the channel images $x_A(t)$, $x_B(t)$, and the common-mode image $u(t)$ along a fixed sample trajectory. If we assume that a synchronization gap exists, we expect that $u(t)$ reveals a consistent digit structure significantly earlier than the resolution of the fine grained details required for pixel level agreement between $x_A(t)$ and $x_B(t)$.

Finally, we perform a targeted causal test. At a chosen reverse timestep $t_{\mathrm{int}}$, we inject noise only into the difference mode
\begin{align}
v(t_{\mathrm{int}})\leftarrow v(t_{\mathrm{int}})+\sigma\xi,\qquad \xi\sim\mathcal{N}(0,I),
\end{align}
leaving $u$ unchanged at that step, and then continue deterministic sampling.
We compare the final outputs with and without intervention under identical initial noise. The existence of the synchronization gap predicts that perturbations within the window where $u$ is stable but $v$ is not,
produce persistent desynchronization effects, whereas perturbations outside the window are largely corrected.

\subsection{Protocol II: Cloning Based Speciation Curves}
\label{subsec:cloning_speciation}
Deterministic traces characterize a single mean reverse trajectory, but the theory of speciation is fundamentally about branching of stochastic paths. We therefore implement a cloning based observation of the speciation time following a similar procedure as in \cite{biroli2024}. In this case, we use DDPM \cite{ho2020denoising} with anisotropic noise for the case of nonzero coupling strength $g$. Since DDPM is much slower in the second protocol, we only use MAIN samples. We also reduce the number of available classes to two for better comparison with theory where we have two clusters.

To implement the standard DDPM reverse updates with unit variance noise, we work in rescaled coordinates
\begin{equation}
y_u = \frac{u}{\sqrt{1-g}},\qquad y_v = \frac{v}{\sqrt{1+g}},
\end{equation}
and perform the usual DDPM step in $y$-space, and then map back to $(u,v)$ before finally using the change of variables map back to $(x_A,x_B)$. 
This yields the correct stochastic reverse kernel for the mode shaped forward law \eqref{eqn:EpsUV}.

We evaluate class identity using swap invariant images derived from the final sample $(x_A,x_B)$ as
\begin{equation}
u_{\mathrm{img}}=\frac{x_A+x_B}{2},\qquad
v_{\mathrm{img}}=\frac{x_A-x_B}{\sqrt{2}}.
\end{equation}
The factor $1/2$ in $u_{\mathrm{img}}$ keeps pixel values in $[0,1]$ and does not affect binary classification. We train two classifiers $C_u$ and $C_v$ on $u_{\mathrm{img}}$ and $v_{\mathrm{img}}$, respectively, with sign invariance enforced by using the symmetrized logits $\tfrac12(C_v(v_{\mathrm{img}})+C_v(-v_{\mathrm{img}}))$. This yields predicted labels $\widehat{y}_u,\widehat{y}_v\in\{0,1\}$.

Fix a set of scan times $\mathcal{T}_{\mathrm{scan}}\subset\{0,\dots,T-1\}$.For each run, we generate a master reverse trajectory and cache the intermediate states $z_t$ for all $t\in\mathcal{T}_{\mathrm{scan}}$.
At each cached time $t$, we create two independent clones that start from the same $z_t$ and continue to $t\to 0$ with independent reverse noise. For each mode $m\in\{u,v\}$ we define the cloning agreement
\begin{equation}
\phi_m(t):=\mathbb{P}\!\left(\widehat{y}^{(1)}_m=\widehat{y}^{(2)}_m\right),
\qquad m\in\{u,v\}.
\end{equation}
Empirically, $\phi_m(t)$ is estimated as an average over batches and independent master trajectories.

Raw agreement scores can be misleadingly high due to classifier bias or marginal class imbalances. To correct for this, we report the baseline corrected agreement
\begin{equation}
\label{eqn:ExcPhi}
\phi^{\mathrm{ex}}_m(t)=\frac{\phi_m(t)-\phi^{\mathrm{indep}}_m}{1-\phi^{\mathrm{indep}}_m},
\end{equation}
where $\phi^{\mathrm{indep}}_m$ is the agreement probability for two independent samples from the model. This maps independent behavior to $0$ and perfect agreement to $1$. For a threshold $\phi^\star=0.55$, we define the speciation time as the earliest reverse step at which the curve $\phi_m(t)$ crosses $\phi^\star$, accounting for the reverse time orientation, this means we have
\begin{equation}
t^{\mathrm{spec}}_m(\phi^\star)
=
\sup\{t\in\mathcal{T}_{\mathrm{scan}}:\ \phi^{\mathrm{ex}}_m(t)\ge \phi^\star\},
\qquad m\in\{u,v\},
\end{equation}
with a linear interpolation between adjacent scan points when reporting non-integer crossing times.
Comparing $t^{\mathrm{spec}}_u$ and $t^{\mathrm{spec}}_v$ provides a reproducible test of the temporal ordering, independent of the deterministic metrics in Protocol~I. To quantify uncertainty, we model the number of agreeing pairs as a binomial count aggregated over repeats and batch size. We compute the $95\%$ Wilson score confidence intervals for the raw agreements $\phi_m(t)$ and since \eqref{eqn:ExcPhi}
is a monotonic transformation, we obtain confidence intervals for the corrected metric $\phi_m^{ex}(t)$ by directly applying the same transformation to the Wilson interval bounds.

\subsection{Results}

\subsubsection*{Protocol I Results}

We demonstrate that the synchronization gap is a robust phenomenon that is significantly diminished in the control setting. We compare samples from our MAIN against a matched CTRL sample which removes semantic pairing while preserving the systems dynamics.

\begin{figure}[ht]
    \centering
    \begin{subfigure}[b]{0.45\textwidth}
    \centering
    \includegraphics[width=\textwidth]{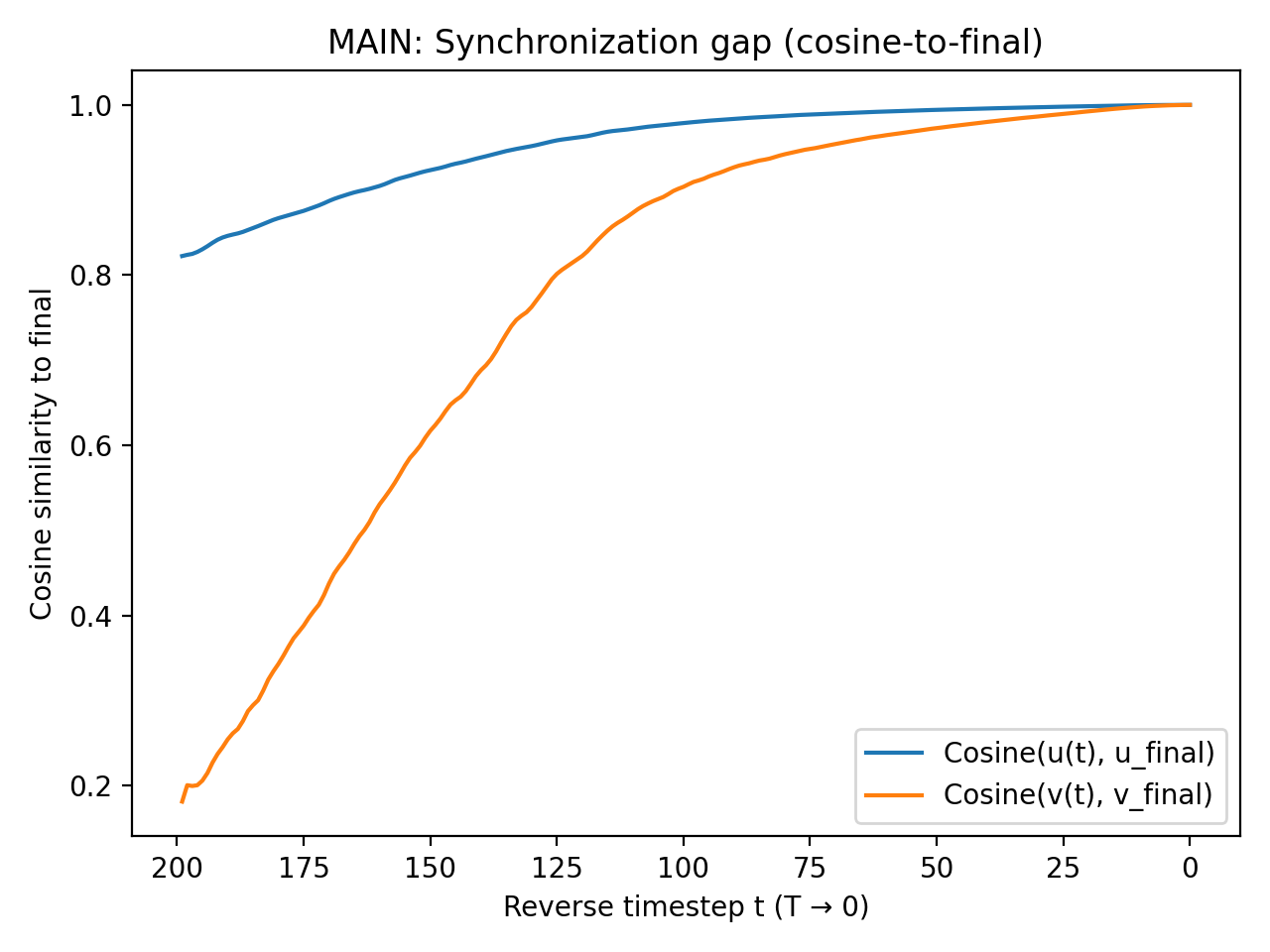}
    \caption{MAIN}
    \end{subfigure}
    \begin{subfigure}[b]{0.45\textwidth}
    \centering
    \includegraphics[width=\textwidth]{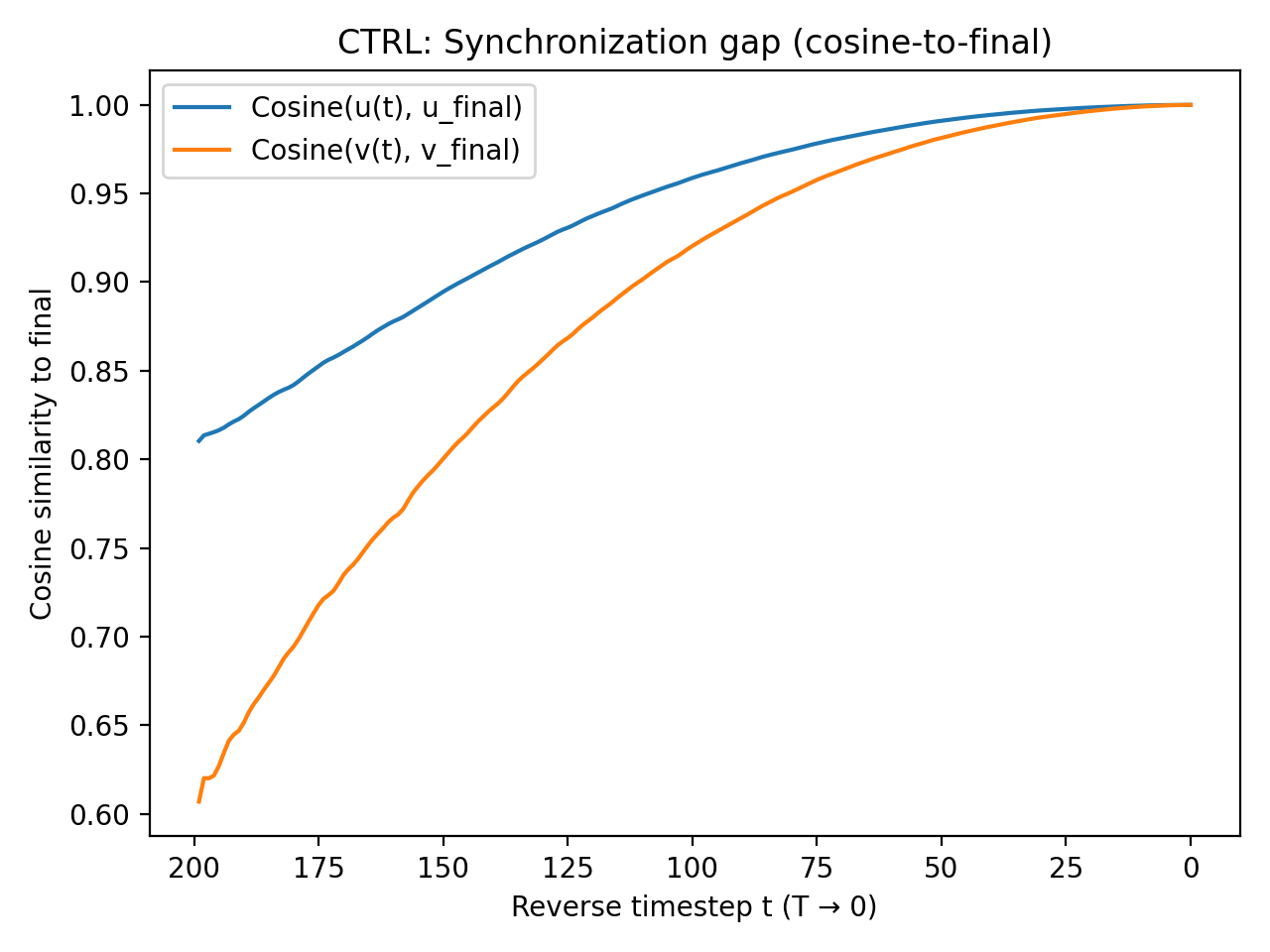}
    \caption{CTRL}
    \end{subfigure}
    \caption{Protocol I synchronization gap. We plot cosine similarity to final metrics for both modes with respect to a reserve timestep. The coupling strength is set to $g=0.3$.}
    \label{fig:SyncGap}
\end{figure}

For each coupling strength $g\in\{0.0,0.3,0.5,0.7\}$, we sweep four random seeds with $T=200$ reverse steps, 20 training epochs, and evaluation batch size 16. \tabref{tab:synch_gap} aggregates the resulting gaps, computed using \eqref{eqn:TimeGaps}. On \figref{fig:SyncGap} we show an example of cosine metric evolution during denoising process for coupling strength $g=0.3$. One could question why there is a gap even for $g=0$, $u$ and 
$v$ have equal noise, but the data geometry makes learning $u$ easier, it retains digit like structure under averaging of correlated pairs in MAIN, while 
$v$ is dominated by transform specific residue. Hence, $u$ aligns with its final state earlier even without dynamical coupling. In the second part of the experiment we will show that indeed for $g=0$ the synchronization gap between speciation times (which we discussed in \secref{sec:IsoCoup} is close to zero).
\begin{table}[ht]
\centering
\small
\setlength{\tabcolsep}{5.0pt}
\begin{tabular}{c|ccc|ccc}
\toprule
& \multicolumn{3}{c|}{MAIN: $\Delta t(\tau)$ (mean $\pm$ sd)} &
  \multicolumn{3}{c}{CTRL: $\Delta t(\tau)$ (mean $\pm$ sd)} \\
$g$ & $\tau=0.90$ & $\tau=0.95$ & $\tau=0.98$ & $\tau=0.90$ & $\tau=0.95$ & $\tau=0.98$ \\
\midrule
0.0 & $-64.0 \pm 15.4$ & $-57.0 \pm 12.1$ & $-43.5 \pm 6.2$ &
      $-25.3 \pm 6.3$ & $-20.0 \pm 4.2$ & $-14.0 \pm 2.2$ \\
0.3 & $-64.5 \pm 6.8$  & $-56.0 \pm 6.1$  & $-49.0 \pm 5.2$ &
      $-29.5 \pm 6.0$ & $-24.3 \pm 3.3$ & $-18.5 \pm 2.6$ \\
0.5 & $-77.2 \pm 29.9$ & $-64.5 \pm 17.5$ & $-53.5 \pm 5.4$ &
      $-31.2 \pm 2.9$ & $-26.5 \pm 3.1$ & $-20.2 \pm 2.2$ \\
0.7 & $-78.0 \pm 26.1$ & $-75.5 \pm 18.7$ & $-61.0 \pm 13.0$ &
      $-36.0 \pm 3.6$ & $-32.8 \pm 3.9$ & $-29.0 \pm 2.2$ \\
\bottomrule
\end{tabular}
\caption{\textbf{Protocol I synchronization gap.}
Aggregate over four seeds with $T=200$ time steps, 20 epochs, and evaluation batch size 16.
Negative $\Delta t(\tau)$ means the $u$-stream reaches the threshold earlier (at higher noise) than the $v$-stream.
Across all $(g,\tau)$, the magnitude of the gap is substantially larger in MAIN than CTRL (typically by a factor $\approx 2$--$3$),
supporting that the effect is not explained by generic per-stream denoising dynamics alone.}
\label{tab:synch_gap}
\end{table}
Two robustness checks preserve the same qualitative conclusion:
(i) increasing the reverse horizon to $T=500$ (e.g.\ $g=0.3$ gives MAIN $\Delta t=-113$ vs.\ CTRL $\Delta t=-72$ at $\tau=0.95$;
$g=0.5$ gives MAIN $\Delta t=-103$ vs.\ CTRL $\Delta t=-76$ at $\tau=0.95$),
and (ii) increasing evaluation batch size (e.g.\ at $g=0.5$, $\tau=0.95$: MAIN $\Delta t=-82$ vs.\ CTRL $\Delta t=-30$).
Thus the synchronization gap is stable across seeds and evaluation settings, while remaining consistently smaller in the control.

\begin{figure}[ht]
    \centering
    \begin{subfigure}[b]{0.45\textwidth}
    \centering
    \includegraphics[width=\textwidth]{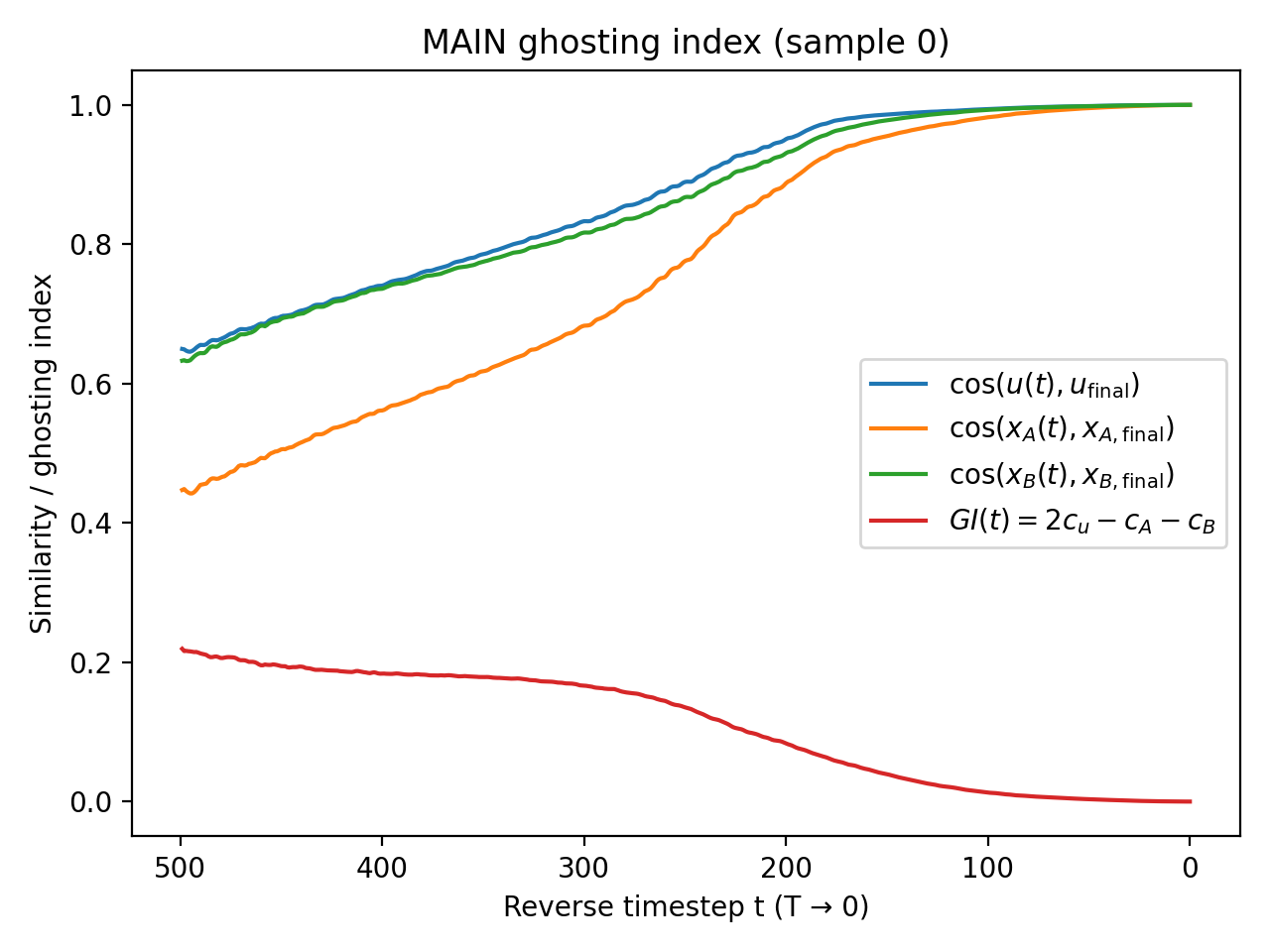}
    \caption{MAIN}
    \label{fig:GhostMain}
    \end{subfigure}
    \begin{subfigure}[b]{0.45\textwidth}
    \centering
    \includegraphics[width=\textwidth]{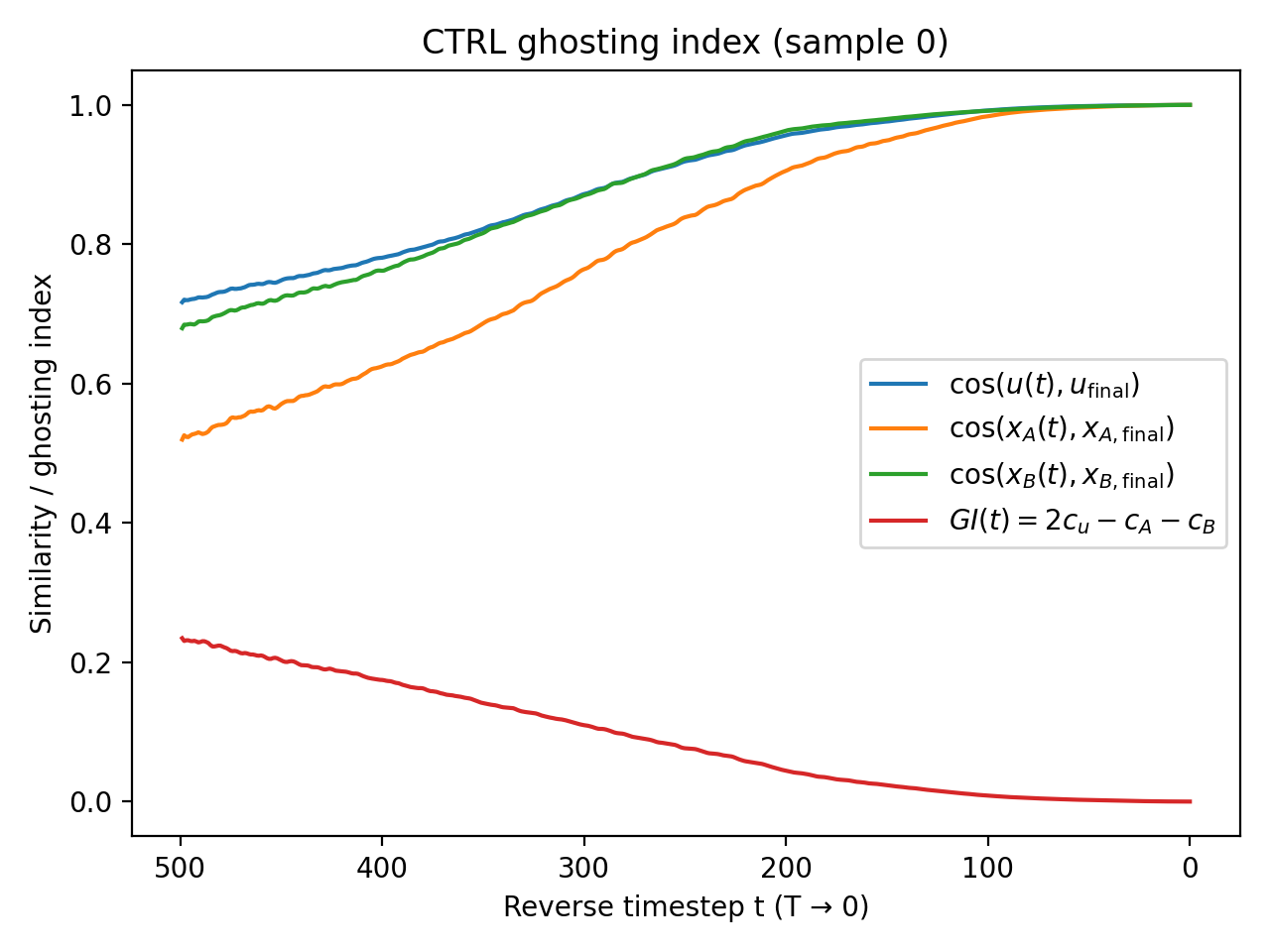}
    \caption{CTRL}
    \label{fig:GhostCtrl}
    \end{subfigure}
    \caption{Protocol I ghosting index. We plot cosine similarity to final frame metrics for common mode $u$ and $x_{A/B}$ as well as ghosting index with respect to reverse timestep. Plots were made for the coupling strength $g=0.3$ and trajectory length $500$.}
    \label{fig:placeholder}
\end{figure}

\begin{figure}[ht]
    \centering
    \includegraphics[width=0.9\linewidth]{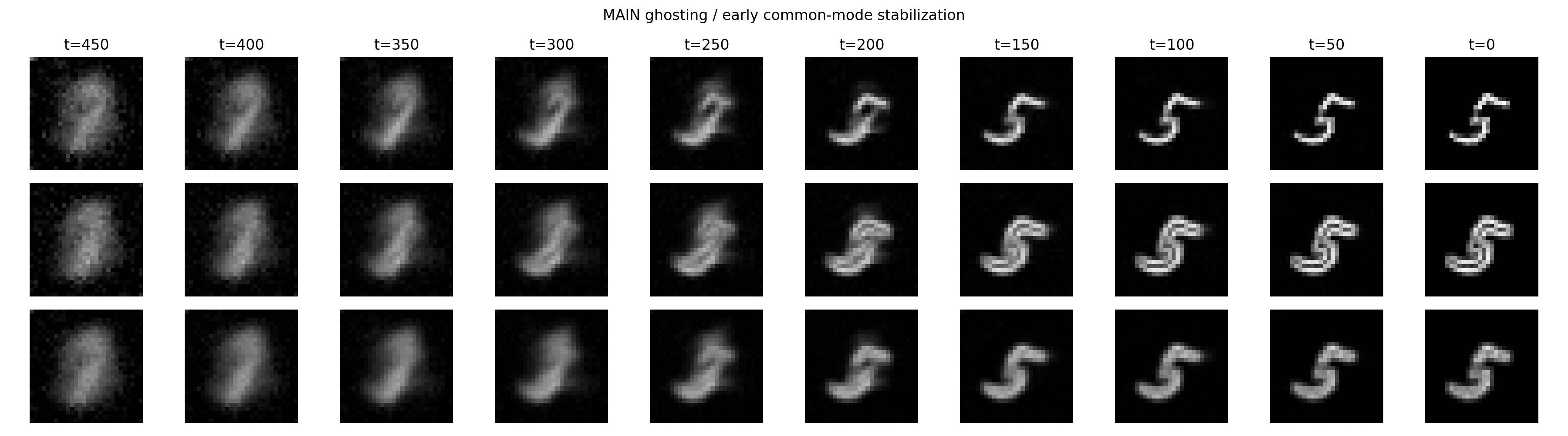}
    \caption{Protocol I generation. We show denoising trajectory for MAIN $g=0.3$, $500$ timesteps. The first row is channel $x_A$, the second is $x_B$ and the third is common channel $u$.}
    \label{fig:MainGhostTraj}
\end{figure}

In the case of Protocol I, ghosting also appears in MAIN and is strongly suppressed in the CTRL sample. To quantify the qualitative ghosting effect, we use cosine similarity for the final frame, we use cosine similarity to the final frame \eqref{eqn:CosSim}:
\begin{equation}
c_u(t) = \cos\big(u(t),u_{\mathrm{final}}\big),\quad
c_A(t) = \cos\big(x_A(t),x_{A,\mathrm{final}}\big),\quad
c_B(t) = \cos\big(x_B(t),x_{B,\mathrm{final}}\big),
\end{equation}
and define the ghosting index
\begin{equation}
GI(t) = 2c_u(t) - c_A(t) - c_B(t).
\end{equation}
In MAIN, $GI(t)$ is steady at high noise and decays toward $0$ as $t\to 0$,
matching the visual observation that $u(t)$ stabilizes into a recognizable structure earlier than either $x_A(t)$ or $x_B(t)$. This is also confirmed by the gap developing between $u$ cosine metric and $x_{A/B}$ cosine metric e.g. on \figref{fig:GhostMain}, where the $u$ cosine metric curve in the middle of denoising process overpasses $x_B$ cosine metric curve.
In CTRL, as shown on \figref{fig:GhostCtrl}, the ghosting index decreases steadily to $0$ and there is no point where $u$ cosine metric curve dominates, consistent with the absence of early common-mode stabilization under the control dynamics. In \figref{fig:MainGhostTraj}, we show the full denoising trajectory to show visually the effect of ghosting. For all three figures, we used $500$ timesteps for better resolution of the effect.


\begin{figure}
    \centering
    \begin{subfigure}[b]{0.43\textwidth}
    \centering
    \includegraphics[width=\textwidth]{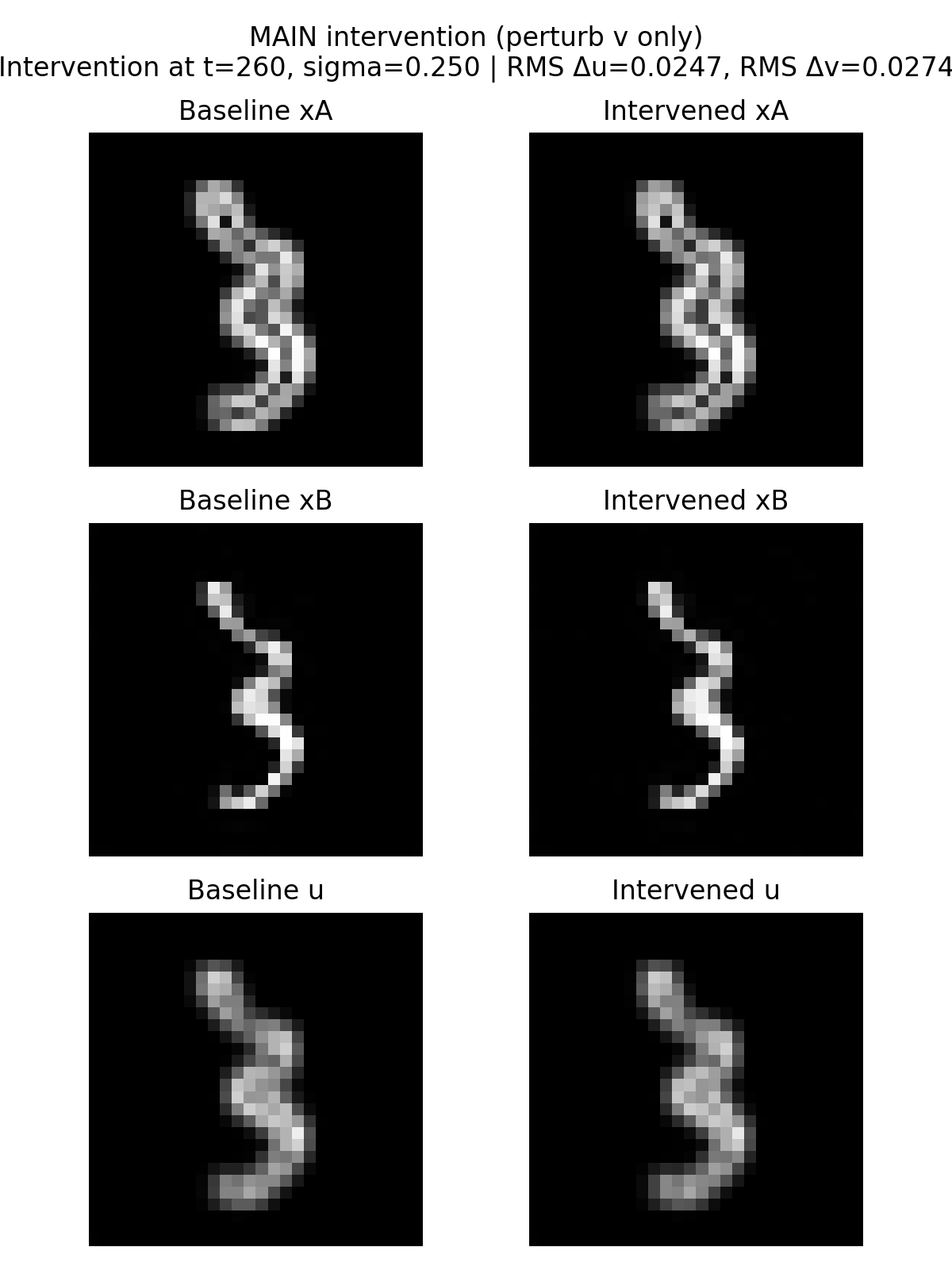}
    \caption{MAIN}
    \end{subfigure}
    \begin{subfigure}[b]{0.43\textwidth}
    \centering
    \includegraphics[width=\textwidth]{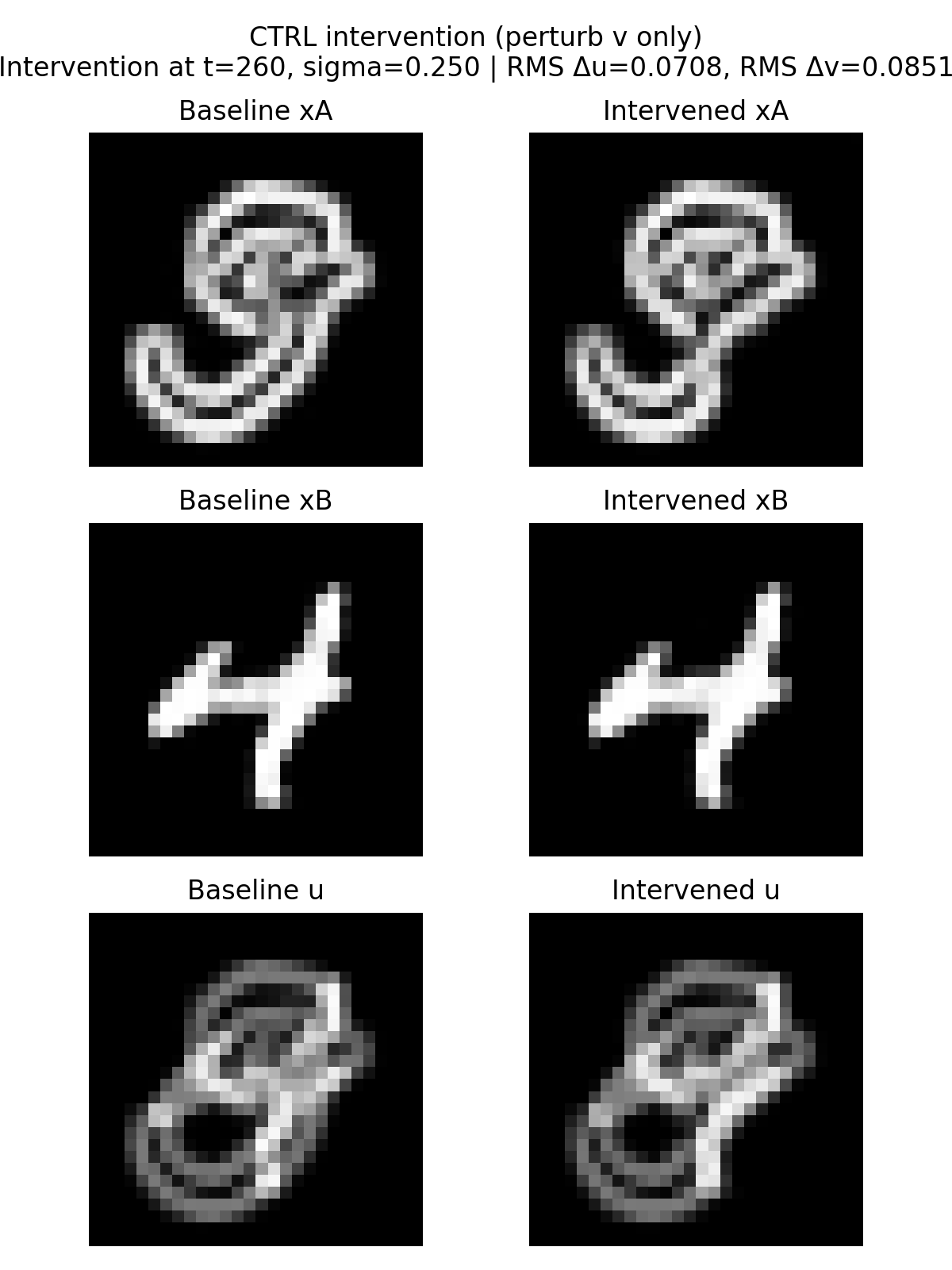}
    \caption{CTRL}
    \end{subfigure}
    \caption{This figure shows the Protocol I intervention. We present intervention plots by injecting noise in $v$ channel and measuring RMS with respect to base values. Plots were made for the coupling strength $g=0.5$ and $500$ timesteps.}
    \label{fig:InterventionPlots}
\end{figure}

We perform an intervention at reverse step $t_{\mathrm{int}}=260$ by perturbing only the $v$-stream with Gaussian noise of scale $\sigma=0.25$,
and measure the resulting RMS changes in the final outputs. An example is shown on \figref{fig:InterventionPlots}.
For CTRL, the perturbation propagates broadly (RMS $\Delta u \approx 0.0708$, RMS $\Delta v \approx 0.0851$),
whereas in MAIN the induced changes are markedly smaller (RMS $\Delta u \approx 0.0247$, RMS $\Delta v \approx 0.0274$).
This is consistent with MAIN exhibiting stronger stabilization of the common component against single-stream perturbations,
in line with the ghosting and synchronization-gap signatures.

\subsubsection*{Protocol II}
\begin{figure}[ht]
    \centering
    \includegraphics[width=0.9\linewidth]{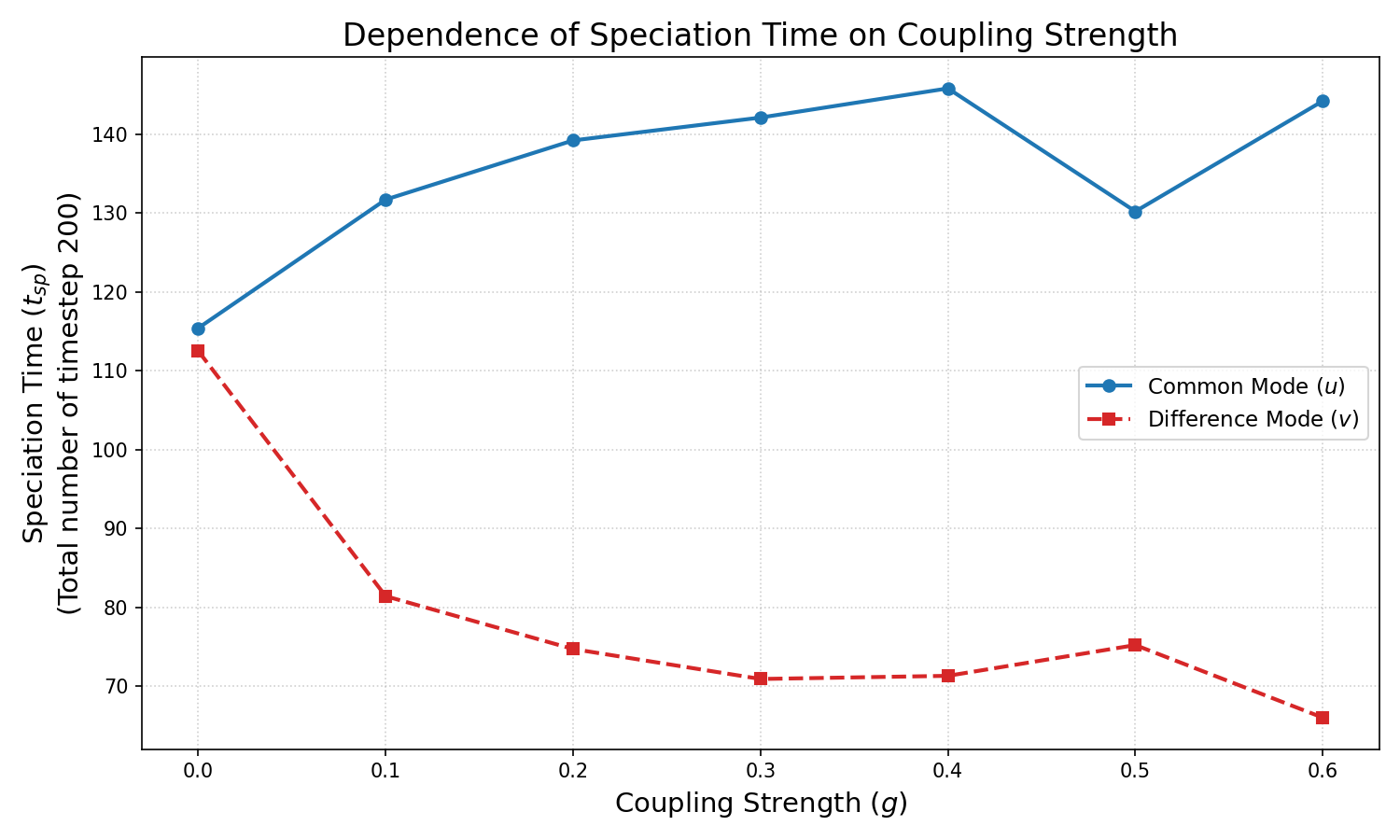}
    \caption{Protocol II synchronization gap. Using cloning procedure we plot common and difference mode dependence on the coupling strength. The appearance of synchronization gap and different scaling of SNRs associated to both modes matches expectations discussed in the theory part.}
    \label{fig:SpeciationSweep}
\end{figure}

\begin{figure}[ht]
    \centering
    \begin{subfigure}[b]{0.43\textwidth}
    \centering
    \includegraphics[width=\textwidth]{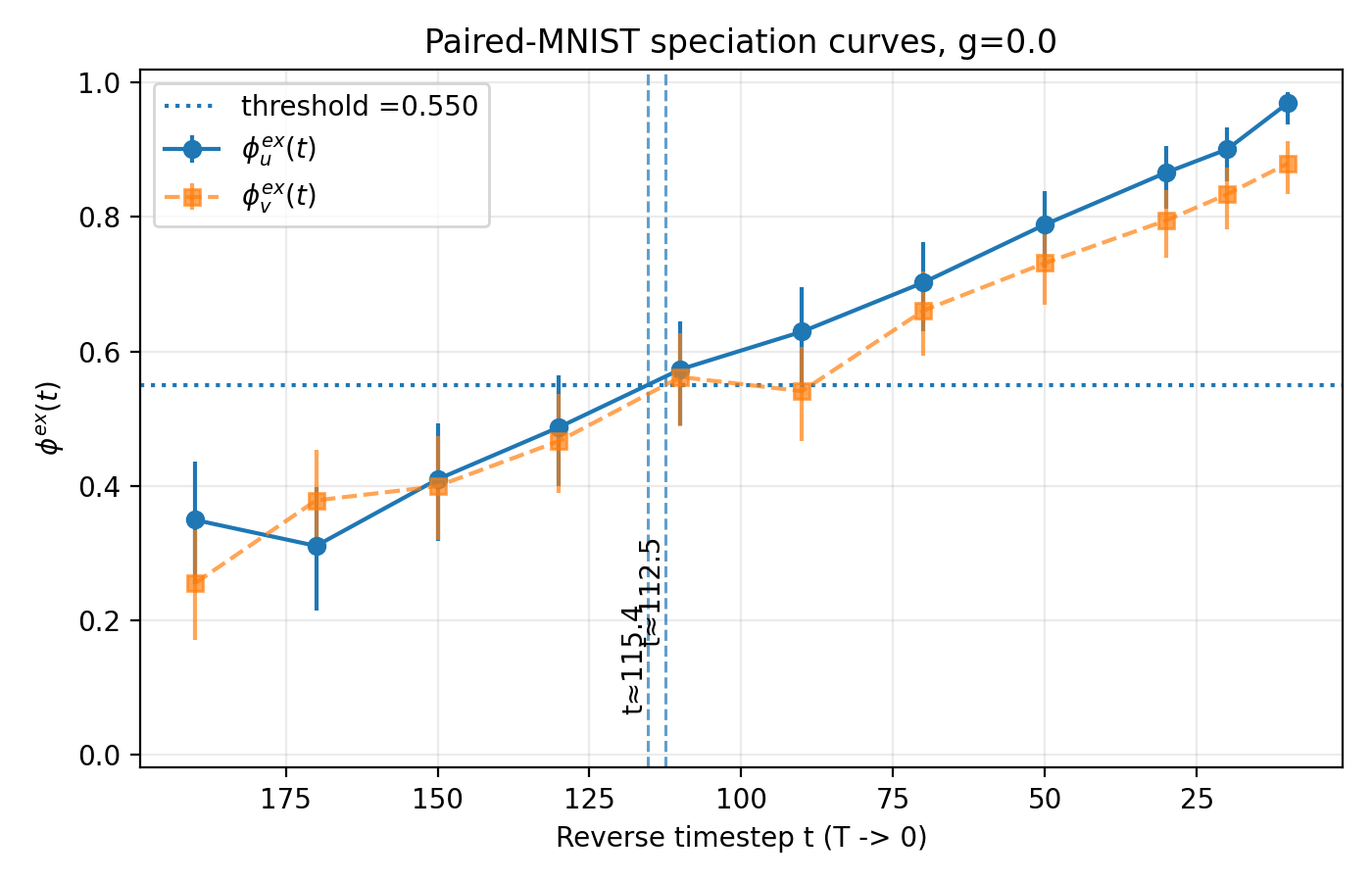}
    \caption{Speciation curves, $g=0.0$.}
    \label{fig:SpecG0}
    \end{subfigure}
    \begin{subfigure}[b]{0.43\textwidth}
    \centering
    \includegraphics[width=\textwidth]{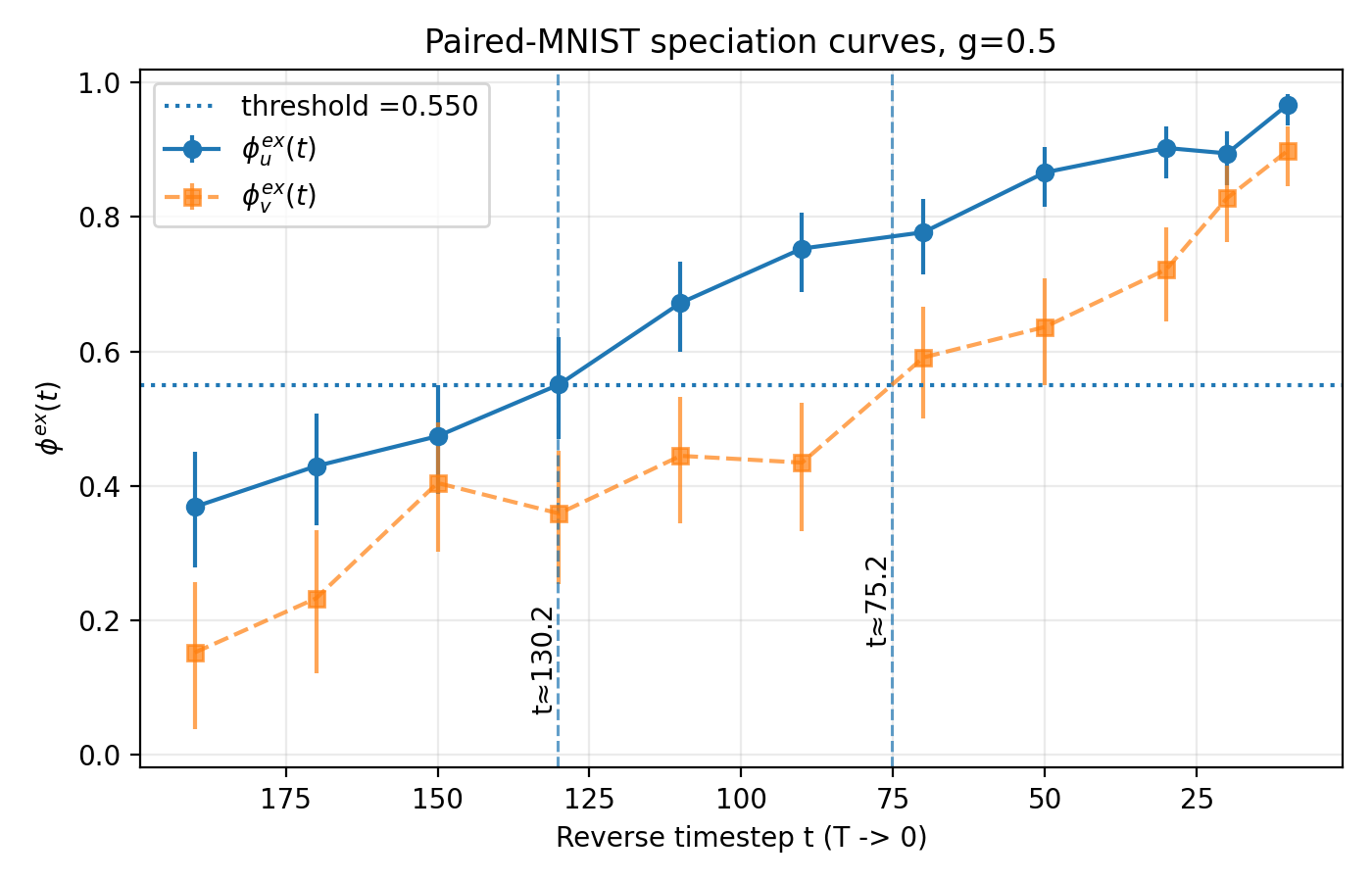}
    \caption{Speciation curves, $g=0.5$.}
    \label{fig:SpecG05}
    \end{subfigure}
    \label{fig:SpeciationExamples}
    \caption{We show evolution of $\phi_u^{\text{ex}}$ and $\phi_v^{\text{ex}}(t)$ during denoising process for coupling strengths $g=0.0$ and $g=0.5$. We record time when they cross the threshold line corresponding to regime transition. For vanishing coupling the gap is almost non-existent.}
\end{figure}

In this case, we directly visualize $\phi^{\mathrm{ex}}_u(t)$ and $\phi^{\mathrm{ex}}_v(t)$ \eqref{eqn:ExcPhi} as functions of reverse time for varying coupling strengths $g$. We summarize an average over 5 trials with 128 batch size for coupling strengths $g\in\{0.0,0.1,0.2,0.3,0.4,0.5,0.6\}$ in \figref{fig:SpeciationSweep}. We also include two representative examples at $g=0.0$ \figref{fig:SpecG0} and $g=0.5$ \figref{fig:SpecG05} to show directly the emergence of the synchronization gap. As mentioned before in the cloning experiment, it is clear that the gap is coupling dependent and it almost disappears for $g=0.0$. Across the sweep, increasing coupling produces larger and more persistent separation between the $u$ and $v$ curves, supporting a genuine coupling-driven synchronization phenomenon rather than a thresholding artifact. We sweep up to $g=0.6$. For larger $g$, the rescaled DDPM coordinates $y_u=u/\sqrt{1-g}$ become numerically ill-conditioned, and empirically the $v$-mode agreement curve often fails to cross $\phi^\star$ reliably within the finite reverse horizon,
making $t^{\mathrm{spec}}_v$ ill-defined. We therefore restrict to a moderate-coupling regime where both modes admit stable, replicable threshold crossings.

Collectively, these results support that synchronization gaps, ghosting, and selective intervention sensitivity are intrinsic signatures of the mode SNR split mechanism implemented here. This is consistent with the qualitative
prediction of \secref{sec:IsoCoup} that separating effective mode SNRs produces a temporal window in which global content stabilizes before fine cross-channel discrepancies.

\section{Discussion and Future Directions}

In this work, we have established a rigorous theoretical framework for multimodal generative dynamics by extending the statistical mechanics of diffusion models to coupled systems. By modeling the interaction between modalities as a coupled Ornstein Uhlenbeck process, we identified the thermodynamic constraints that govern the simultaneous emergence of content coherence and cross modal alignment. 

Our central theoretical contribution is the analytic derivation of speciation and collapse time for coupled diffusion model. In the symmetric coupling case we found the synchronization gap, a temporal window where the common eigenmode has transitioned into the speciation regime, while the difference eigenmode  remains in the noise regime. In the anisotropic coupling case, we showed a dependence of speciation time on both coupling strength as well the alignment angle between the conditioning signal and the target mean.

We checked these theoretical predictions through extensive experimentation using standard diffusion model architectures the for symmetric coupling case and the exact score experiment for anisotropic coupling. We found the confirmation of the existence of the synchronization gap in a baseline dataset and showed its width dependence on the coupling strength. We also showed that different coupling strengths and schedules could improve misaligned generative processes.

The implications of this framework extend beyond curiosity, offering actionable insights for the design of next generation multimodal diffusion architectures. Our findings advocate for adaptive coupling schedules that can accelerate speciation time without significant modifications to the collapse time. One also needs to consider the synchronization gap so that the model does not generate samples until both modalities stabilized into II regime.

Several promising avenues for future research emerge from this study. While our analysis focused on dealing with the data space directly, it is a natural question to try to investigate how the synchronization gap manifests under more recent architecture designs like those in Latent Diffusion Models \cite{Rombach2021-li}  where the natural space to do the analysis in is the latent space.

Our current model assumes a static coupling strength. Future work could explore time dependent couplings $M(t)$ which could admit derivative expansions introducing additional time scales during the generation. Additionally, this framework would allow us to begin designing coupling schedules analogous to the noise schedules that we use for $\Sigma_W$. 

Formally, this extends our framework to a non-autonomous stochastic system, modulo the noise term, of the form $d Z(t)=M(t) Z(t) d t$, we could begin to understand the structure of the time dependent case by integrating iteratively as we do with Dyson series, the result being
\begin{equation}
Z(t)=\left(I+\int_0^t M\left(s_1\right) d s_1+\int_0^t \int_0^{s_1} M\left(s_1\right) M\left(s_2\right) d s_2 d s_1+\ldots\right) Z(0),
\end{equation}
which is effectively the signature transform \cite{Chen1977IteratedPI}. The importance of this results lies in the fact that the signature transform encodes the noncommutativity of the underlying stochastic paths through the tensor algebra \cite{Lyons1994DIFFERENTIALED} which reveals that the time dependent coupling case imposes a specific chronological ordering on the information flow. The higher order terms of the signature should effectively capture the sequence in which the modalities interact, suggesting that the order of the cross modal diffusion is as critical as the coupling strength itself. The use of the signature transform in deep learning has been previously leveraged and is reviewed in \cite{Chevyrev_2025}. 

Finally, we would like to point out that although the analysis has shown that coupled OU oscillators are incredibly useful in predicting several distinct phenomena in diffusion models, it is not necessarily a fundamental fact that all diffusion models would necessary fall into this linear description. We argued that the key property was that the SDE was mean reverting such that no numerical divergence appear. Let $X_t \in \mathbb{R}^d$ and $x$ its realization, a natural nonlinear generalization would therefore be a cubic mean reverting SDE with additive noise of the form
\begin{equation}
d X_t=\left(a X_t-b X_t^3\right) d t+\sigma d W_t, \quad b>0, \sigma>0, a \in \mathbb{R}
\end{equation}
For such an SDE, we would still be able to find a stationary distribution for which we could condition the reverse process to, the Fokker Planck equation can be directly solved to give the Gibbs measure
\begin{equation}
\pi(x) \propto \exp \left(\frac{\alpha}{2}\|x\|^2-\frac{\beta}{4}\|x\|^4\right),
\end{equation}
where $Z$ is a normalization factor.  In this case, the theoretical analysis suggests that we can no longer simply train a model with a nonlinear forward process and then take a simple Gaussian noise for the reverse process. The generation process would simply always output noise. From a physical point of view, it is now clear that the analysis of such systems can be aided by the techniques of nonequilibrium field theory \cite{zinn2021quantum}. We expect this to be an eventual necessary tool for the analysis of nonlinear theoretical constructions for diffusion models due to three key observations. Having nonlinear interaction terms allows us to begin modeling nonlinear coupling architectures such as those that appear in cross attention \cite{vaswani2017attention} which is based on a nonlinear state dependent coupling i.e., the attention mechanism. Second, making the connection between nonlinear stochastic systems and nonequilibrium field theory naturally gives us an analytic method of trying to design coupling schedules for the parameters $a$ and $b$ through standard renormalization group flow arguments \cite{wilson1975renormalization}.
Lastly, finding stationary distributions from which theoretical analysis can begin is one of the shining points of quantum field theory, in the simple example of the Gibbs measure, one can easily show that the prior is now bimodal noise in the statistics sense of the word modality. This effectively shifts our generative priors from a maximal entropy Gaussian to a Boltzmann distribution.

Finally, testing these dynamics on large scale, heterogeneous datasets e.g. video-audio-text triplets is a natural extension. We hypothesize that as the number of modalities increases, the spectral gap between the fastest and slowest eigenmodes will widen, exacerbating desynchronization artifacts. Verifying the scaling laws of the collapse time $t_C$ in these hyper multimodal regimes will be crucial for the stability of future foundation models.

\section*{Acknowledgments}
We would like to thank Ori Ganor, David T. Limmer, Wojciech Musial, Edward Yam and, Viola Zixin Zhao for helpful discussions surrounding related nonequilibrium systems, machine learning and possible multi-agent reinforcement learning followups to this work.  This work has been supported by the Leinweber Institute for Theoretical Physics at UC Berkeley.

\appendix

\section{Derivation Of The Cumulant Generating Function For Symmetric Coupling}
\label{app:DerCGF}

In this appendix, we derive the asymptotic limit of the Cumulant Generating Function (CGF) stated in \eqref{eqn:LambdaSym}. We begin with the finite-dimensional definition in \eqref{eqn:CumulantGen} and utilize the diagonalized basis introduced in \secref{sec:IsoCoup}. If we denote 
\begin{align}
    q_{\pm}=\sigma_W^2\frac{e^{2\lambda_\pm t}-1}{2\lambda_\pm},
\end{align}
then the reduced energy of \eqref{eqn:CumulantGen} can be decomposed as
\begin{align}
    \varepsilon_t(\tilde{z},z)=\varepsilon_{+,t}(\tilde{z}_+,z_+)+\varepsilon_{-,t}(\tilde{z}_-,z_-), \qquad \varepsilon_{\pm,t}(\tilde{z},z)=\frac{1}{4dq_\pm(t)}||\tilde{z}_\pm-e^{\lambda_\pm t}z_\pm||^2,
\end{align}
where $z_\pm$ is a decomposition of $z$ on the diagonal basis. In the random energy model for a fixed point $\tilde{z}$, the energies $\{\varepsilon_t\}_{i=1}^n$ are treated as i.i.d. random variables. Therefore, we can write a conditional CGF
\begin{align}
    \Lambda_t(\beta; \tilde{z})=\lim_{d\rightarrow\infty}\frac{1}{2d}\log\EV_z[\exp(-\beta 2d\varepsilon(\tilde{z},z))],
\end{align}
where $z\sim \CN(0, \sigma^2I_{2d})$. The Strong Law of Large Numbers implies that the normalized squared norms converge almost surely and we get
\begin{align}
\label{eqn:LLN}
    \frac{1}{d}\|\tilde{z}_\pm\|^2 \xrightarrow[d\to\infty]{a.s.} c_\pm(t).
\end{align}
Since $\Lambda_{t}(\beta; \tilde{z})$ is a continuous function of the sufficient statistics $\frac{1}{d}\|\tilde{z}_\pm\|^2$, the Continuous Mapping Theorem ensures that the random variable $\Lambda_t(\beta; \tilde{z})$ converges almost surely to a deterministic limit $\Lambda_t(\beta)$. 

Since the energy functional decomposes into independent terms for the $+$ and $-$ modes, the expectation factorizes. We may therefore compute the contribution of a single mode for a single mode and hence we drop the $\pm$ from the notation temporarily.

Denoting $a=e^{2\lambda t}$, using the standard identity for the expectation of a quadratic exponential under a Gaussian measure $z \sim \mathcal{N}(0, \sigma^2 I_d)$, we have
\begin{align}
\mathbb{E}_z\left[e^{-\frac12 z^\intercal A z + b^\intercal z}\right]
  =
  \det(I_d+\sigma^2 A)^{-1/2}
  \exp\left(\frac{\sigma^2}{2}b^\intercal (I_d+\sigma^2 A)^{-1}b\right),
\end{align}
where we used the fact that $z_\pm\in \mathbb{R}^d$, we find
\begin{align}
\mathbb{E}_z\left[\exp\!\Big(-\frac{\beta}{2q}\|\tilde{z}-a z\|^2\Big)\right]
  =\
  \Big(1+\beta\,\frac{\sigma^2 a^2}{q}\Big)^{-d/2}\,
  \exp\left(
    -\frac{\beta}{2q}\,\frac{1}{1+\beta\,\frac{\sigma^2 a^2}{q}}\,\|\tilde z\|^2
  \right).   
\end{align}
We can use \eqref{eqn:Chi} to write
\begin{align}
    \beta\,\frac{\sigma^2 a^2}{q}=\beta\chi_\pm, 
\end{align}
hence 
\begin{align}
  \Lambda_{\pm,t}(\beta)
  &=
  \lim_{d\to\infty}\frac{1}{2d}\log
  \mathbb{E}_{z_\pm}\!\left[\exp\!\Big(-\beta\,2d\,\widehat\varepsilon_{\pm,t}(\tilde{z}_\pm,z_\pm)\Big)\right]
  \nonumber\\
  &=
  \lim_{d\to\infty}\frac{1}{2d}\left[
    -\frac{d}{2}\log(1+\beta\chi_\pm)
    -\frac{\beta}{2q_\pm}\frac{1}{1+\beta\chi_\pm}\|\tilde{z}_\pm\|^2
  \right]
  \nonumber\\
  &=
  -\frac14\log(1+\beta\chi_\pm)
  -\frac{\beta}{4}\,\frac{1+\chi_\pm}{1+\beta\chi_\pm}.
  \label{eq:B_Lambda_mode},
\end{align}
using \eqref{eqn:LLN}. Summing over both $+$ and $-$ modes, we recover \eqref{eqn:LambdaSym}.

\section{Explicit Form Of Diffusion Kernel For Anisotropic Coupling}
\label{app:ExpCt}

In this appendix, we derive the closed form expressions for the covariance matrix $C(t)$ and the diffusion kernel $K(t)$ in the asymmetric coupling regime. We consider the process governed by the drift matrix $M \in \mathbb{R}^{2\times 2}$ given by
\begin{align}
    M = \begin{pmatrix} -\beta & 0 \\ g & -\beta \end{pmatrix}.
\end{align}
The associated matrix exponential is $e^{M t} = e^{-\beta t} \begin{pmatrix} 1 & 0 \\ gt & 1 \end{pmatrix}$ which can be computed via the nilpotency identity mentioned in \secref{sec:theory}.

The noise covariance matrix $Q(t)$ is defined by the integral $Q(t) = \sigma_W^2 \int_0^t e^{M s} (e^{M s})^\top ds$. To facilitate the integration, we define the following auxiliary functions
\begin{align}
    u(t) &= 1-e^{-2\beta t}, \\
    k(t) &= 1-e^{-2\beta t}(1+2\beta t), \\
    h(t) &= 1-e^{-2\beta t}(1+2\beta t+2\beta^2t^2).
\end{align}
Computing the elementwise integrals of the matrix exponential yields the block structure $Q(t) = \begin{pmatrix} q_{11} & q_{12} \\ q_{12} & q_{22} \end{pmatrix} \otimes I_d$, with components
\begin{align}
    q_{11}(t) &= \sigma_W^2\frac{u(t)}{2\beta}, \\
    q_{12}(t) &= \sigma_W^2\frac{g k(t)}{4\beta^2}, \\
    q_{22}(t) &= \sigma_W^2\left(\frac{u(t)}{2\beta}+\frac{g^2 h(t)}{4\beta^3}\right).
\end{align}
The total state covariance $C(t)$ evolves according to $C(t) = e^{Mt} C(0) e^{M^\top t} + Q(t)$. Assuming an initial diagonal covariance $C(0) = \text{diag}(\sigma_x^2, \sigma_y^2) \otimes I_d$, we obtain the components
\begin{align}
    C_{11}(t) &= e^{-2\beta t}\sigma_x^2 + q_{11}(t), \\
    C_{12}(t) &= e^{-2\beta t}gt\sigma_x^2 + q_{12}(t), \\
    C_{22}(t) &= e^{-2\beta t}\sigma_y^2 + e^{-2\beta t}g^2t^2\sigma_x^2 + q_{22}(t).
\end{align}

We now compute the auxiliary kernel defined as $K(t) := C(t)^{-1} (M + \sigma_W^2 C(t)^{-1})^{-1} C(t)^{-1}$. Let $\Delta(t) = \det(C(t)) = C_{11}C_{22} - C_{12}^2$. The term involving the drift is given by
\begin{align}
     M + \sigma_W^2 C(t)^{-1} = \begin{pmatrix}
        -\beta + \frac{\sigma_W^2 C_{22}}{\Delta} & -\frac{\sigma_W^2 C_{12}}{\Delta} \\
        g - \frac{\sigma_W^2 C_{12}}{\Delta} & -\beta + \frac{\sigma_W^2 C_{11}}{\Delta}
    \end{pmatrix}.
\end{align}
The determinant of this matrix simplifies to
\begin{align}
    \det (M + \sigma_W^2 C(t)^{-1}) = \frac{D(t)}{\Delta(t)},
\end{align}
where we have defined the quantity $D(t)$ as
\begin{align}
    D(t) := \beta^2 \Delta(t) - \beta \sigma_W^2 (C_{11}(t) + C_{22}(t)) + g \sigma_W^2 C_{12}(t) + \sigma_W^4.
\end{align}
Finally, by applying the standard inversion formula for $2\times 2$ block matrices, the kernel $K(t)$ takes the form
\begin{align}
    K(t) = \frac{1}{\Delta(t) D(t)} \begin{pmatrix} N_{11}(t) & N_{12}(t) \\ N_{21}(t) & N_{22}(t) \end{pmatrix}.
\end{align}
Algebraic manipulation yields the following expressions for the numerators
\begin{align}
    N_{11}(t) &= \sigma_W^2 C_{22}(t) - \beta \left(C_{22}(t)^2 + C_{12}(t)^2\right) + g C_{12}(t)C_{22}(t), \\
    N_{12}(t) &= C_{12}(t) \left[ \beta \left(C_{11}(t) + C_{22}(t)\right) - g C_{12}(t) - \sigma_W^2 \right], \\
    N_{21}(t) &= \beta C_{12}(t)\left(C_{11}(t) + C_{22}(t)\right) - \sigma_W^2 C_{12}(t) - g C_{11}(t)C_{22}(t), \\
    N_{22}(t) &= \sigma_W^2 C_{11}(t) - \beta \left(C_{11}(t)^2 + C_{12}(t)^2\right) + g C_{11}(t)C_{12}(t).
\end{align}

\section{Exact-score OU Toy Experiment For Anisotropic Relaxation Matrix}
\label{sec:toy_ou_experiment}

We construct a minimal conditional generation experiment for anisotropic coupling discussed in \secref{sec:AsymCoup} in order to observe the effect of coupling strength on the model. For simplicity, in the experiment, we use the reverse time sampler with an exact (population) conditional score. This removes the need for neural network to learn the score function and isolates the dynamical effect. 

Let us first briefly describe experimental setup. We work in dimension $d=32$. The data distribution is a two-component Gaussian mixture for both source $X$ and target $Y$:
\begin{align}
x_0 = s\,\mu_x + \sigma_{\text{data}} \varepsilon_x,\quad
y_0 = s\,\mu_y + \sigma_{\text{data}} \varepsilon_y,
\end{align}
where $\sigma_{\text{data}}=1$, $\varepsilon_x,\varepsilon_y \sim \mathcal N(0, I_d)$, and the class label $s$ is sampled uniformly from $\{+1, -1\}$. 

To model geometric misalignment, we confine the mean vectors to a 2D signal plane spanned by deterministic orthonormal basis vectors $u, v \in \mathbb{R}^d$. We fix the signal-to-noise ratio such that the per-dimension mean squared amplitude is $m^2$. The means are parameterized by a relative angle $\theta$
\begin{align}
\mu_x = \sqrt{d}\,m\,u, \quad \mu_y = \sqrt{d}\,m\,(\cos\theta\,u + \sin\theta\,v).
\end{align}
We sweep $\theta \in [0, \pi]$ to transition from perfectly aligned domains ($\theta=0$) to orthogonal ($\theta=\pi/2$) and anti-correlated ($\theta=\pi$) domains and average metrics over 2000 Monte Carlo trials per $\theta$.

Using \eqref{eqn:SDE} and \eqref{eqn:AsymMandS} we write asymmetric coupling OU process equations. In this case $X(t)$ does not depend on $Y(t)$, while $Y(t)$ is driven by $X(t)$ through a scalar coupling $g(t)$,
\begin{align}
dX(t) &= -\beta X(t)\,dt + \sigma_W\,dW_x(t), \\
dY(t) &= \bigl(-\beta Y(t) + g(t)\,X(t)\bigr)\,dt + \sigma_W\,dW_y(t), \quad t\in[0,T].
\end{align}
In numerical experiment we set $\beta=1$, $\sigma_W^2=2$, and $T=2$. The first and second moments ($\mu(t), C(t)$) are computed exactly by integrating the corresponding linear ODEs.

Conditioned on a forward path $X_{0:T}$, the exact conditional distribution $P_t(y|x)$ at any time $t$ remains a Gaussian mixture. Using the Schur complement of the joint covariance matrix $C(t)$, the conditional score $\nabla_y \log P_t(y|x)$ is derived analytically. The conditional density is given by:
\begin{align}
P_t(y|x) = \sum_{k \in \{+,-\}} w_k(x,t)\,\mathcal N\bigl(y; m_k(x,t),\,C_{y|x}(t)I_d\bigr),
\end{align}
where the component means $m_\pm$ shift based on the observed $x$, and the weights $w_\pm$ correspond to the posterior class probability given $x$.

We simulate the reverse process $Y_t$ from $T \to 0$ using the Euler Maruyama method with $N=800$ steps. The reverse SDE includes the coupling drift term consistent with the forward process:
\begin{align}
d\widetilde{Y}_t = \Bigl(-\beta \widetilde{Y}_t + g(t)\,X_t - \sigma_W^2\,\nabla_y \log P_t(\widetilde{Y}_t|X_t)\Bigr)\,dt
+ \sigma_W\,d\widetilde{W}_{y,t}.
\end{align}

We compare three simple schedules for coupling constant (defined over forward time $t\in[0,T]$):
\begin{align}
g_{\text{const}}(t)&=g_0,\qquad
g_{\text{late}}(t)=g_0\,\mathbf 1\{t\le t_0\},\qquad
g_{\text{early}}(t)=g_0\,\mathbf 1\{t\ge t_0\}, 
\end{align}
and sweep different coupling magnitudes $g_0\in\{0.2,0.5,1.0\}$.

To determine the effect of coupling and its schedule we plot at $t=0$ observables and their difference wrt. the uncoupled baseline $g_0=0$. We report alignment accuracy, signal MSE, and negative likelihood. In details
\begin{itemize}
    \item For alignment accuracy, we determine the sign of $x_0$ data point $s(x_0)=\text{sgn}(\langle x_0,\mu_x\rangle)$ and compare it to the sign of generated point $s(\tilde{y}_0)=\text{sgn}(\langle \tilde{y}_0, \mu_y\rangle)$. We define the accuracy as a sample average of matching signs. 
    \item Next, we compute MSE 
    \begin{align}
        \frac{1}{2}\EV\left[||(\tilde{y}_0)-s(x_0)(\mu_y)||^2\right].
    \end{align}
    \item We also calculate negative likelihood at $t=0$
    \begin{align}
        \text{NLL}=\EV[-\log P_0(\tilde{y}_0|x_0)].
    \end{align}
\end{itemize}

In all registered metrics, we observe a clear regime transition as $\theta$ increases. For small angles (highly aligned modalities) coupling causes alignment accuracy decrease and signal-subspace MSE increase relative to baseline $g_0=0$. For large angles (misaligned modalities) coupling produces alignment accuracy increase and signal-subspace MSE decrease, with a sign change near $\theta\approx \pi/2$--$2\pi/3$ depending on the observable.

Scheduling strongly modulates this trade-off. Constant coupling is typically the most harmful schedule in the aligned regime ($\theta<\pi/2$), while late coupling has the least detrimental effect for aligned regime yet introduces a boost effect for misaligned one ($\theta>\pi/2$). The conditional NLL illustrates an over-guidance-like effect at large coupling: for $g_0=1.0$ the sampler consistently worsens $\mathrm{NLL}_0$ under a fixed discretization budget, whereas for moderate coupling ($g_0=0.5$) scheduled coupling can achieve near-neutral or improved $\mathrm{NLL}_0$ at large $\theta$. For weak coupling ($g_0=0.2$) the NLL penalty disappears and improvements at large $\theta$ are visible across schedules, albeit with smaller alignment gains.

Overall, this exact-score experiment confirms that cross-modal coupling is not uniformly beneficial: it induces a genuine phase structure controlled by the angle $\theta$, and scheduling provides a practical handle to leverage performance while limiting distributional degradation under realistic sampling budgets.

\begin{figure}[ht]
    \centering
    \begin{subfigure}[b]{0.32\textwidth}
    \centering
    \includegraphics[width=\textwidth]{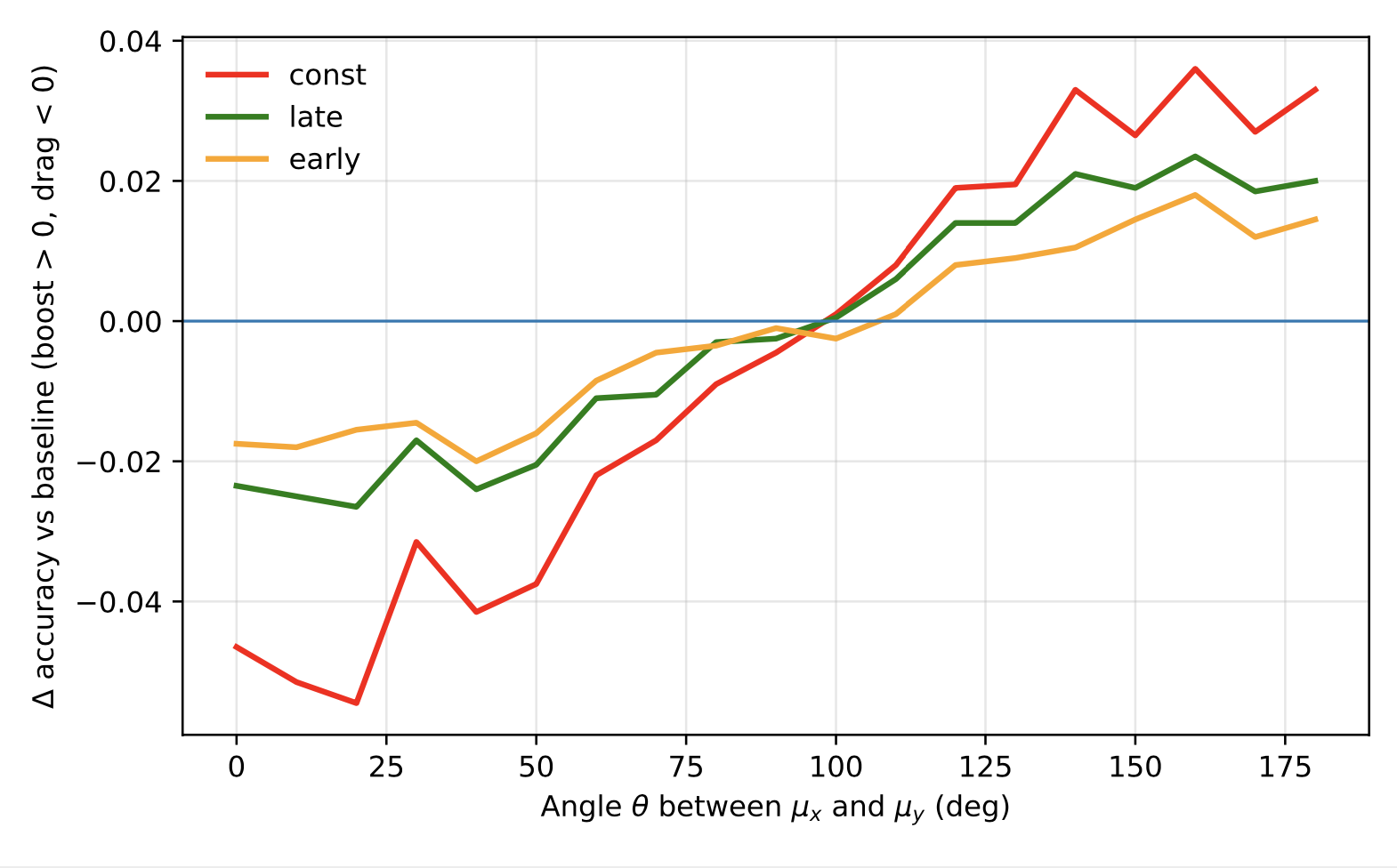}
    \caption{$\Delta$ accuracy vs to baseline.}
    \end{subfigure}
    \centering
    \begin{subfigure}[b]{0.32\textwidth}
    \centering
    \includegraphics[width=\textwidth]{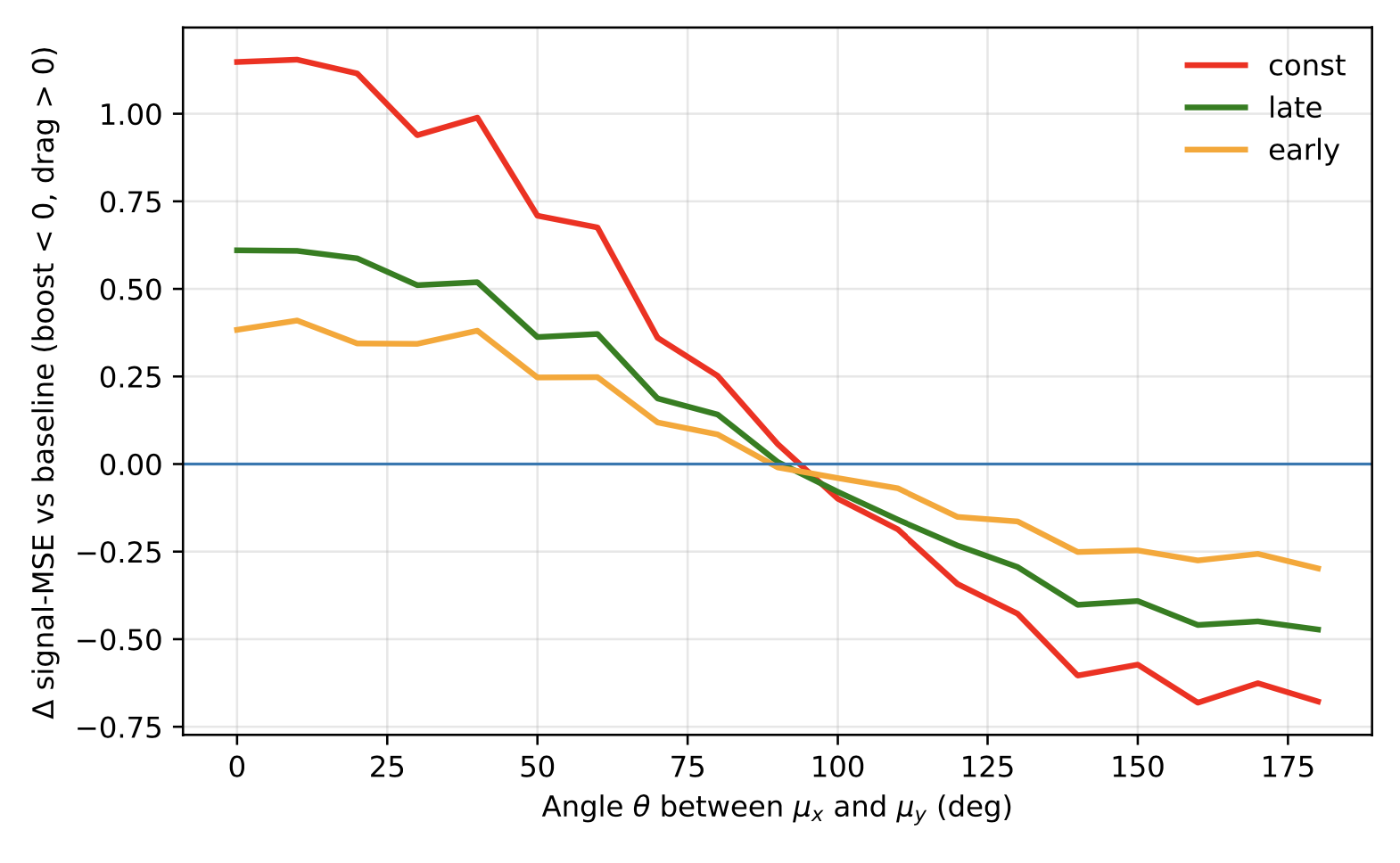}
    \caption{$\Delta$ MSE vs to baseline.}
    \end{subfigure}
    \centering
    \begin{subfigure}[b]{0.32\textwidth}
    \centering
    \includegraphics[width=\textwidth]{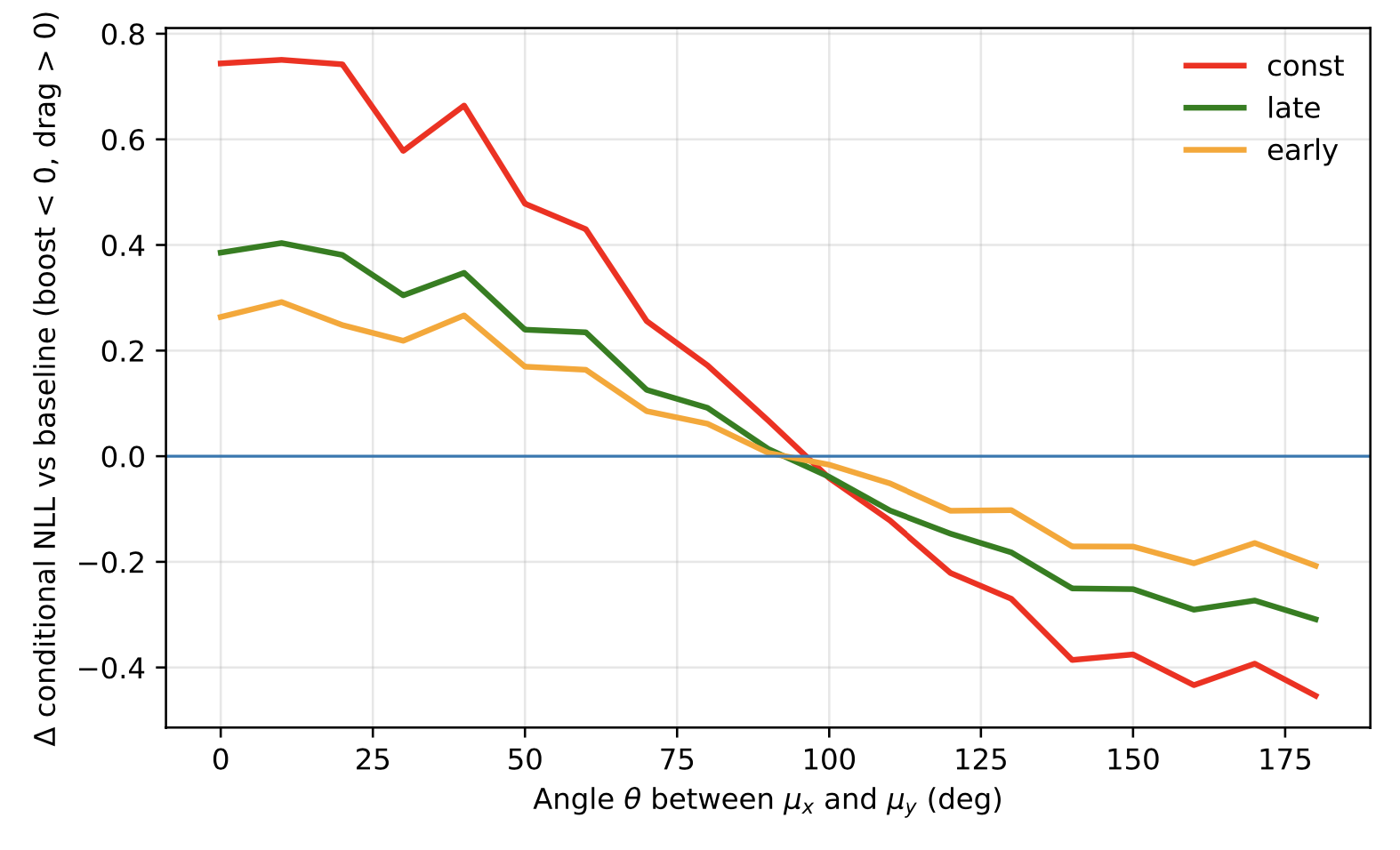}
    \caption{$\Delta$ NLL vs to baseline.}
    \end{subfigure}
    \caption{Results of the OU experiment: $\Delta$ of observable wrt. to baseline for a coupling strength $g_0=0.2$}
\end{figure}
\begin{figure}[ht]
    \centering
    \begin{subfigure}[b]{0.32\textwidth}
    \centering
    \includegraphics[width=\textwidth]{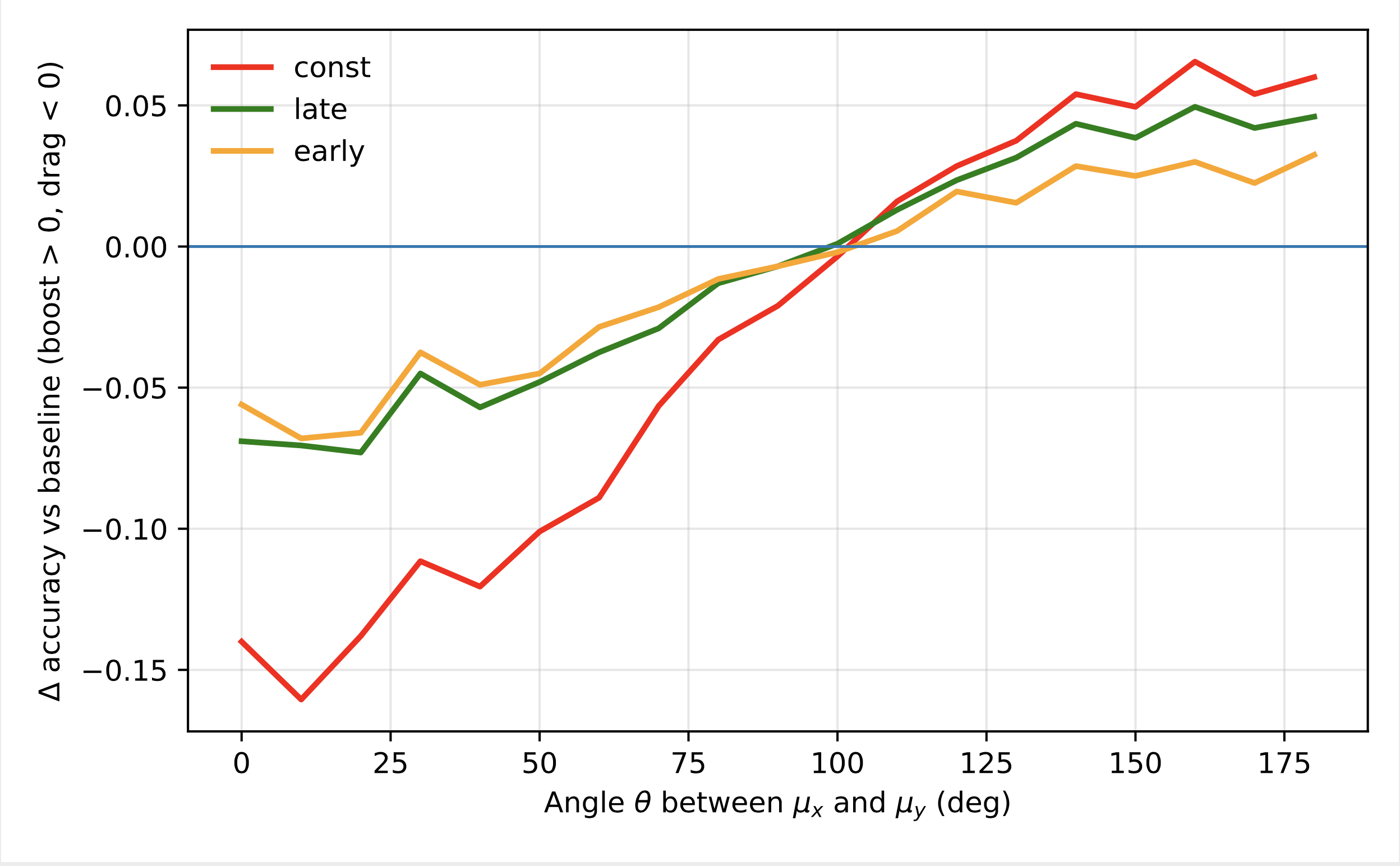}
    \caption{$\Delta$ accuracy vs to baseline.}
    \end{subfigure}
    \centering
    \begin{subfigure}[b]{0.32\textwidth}
    \centering
    \includegraphics[width=\textwidth]{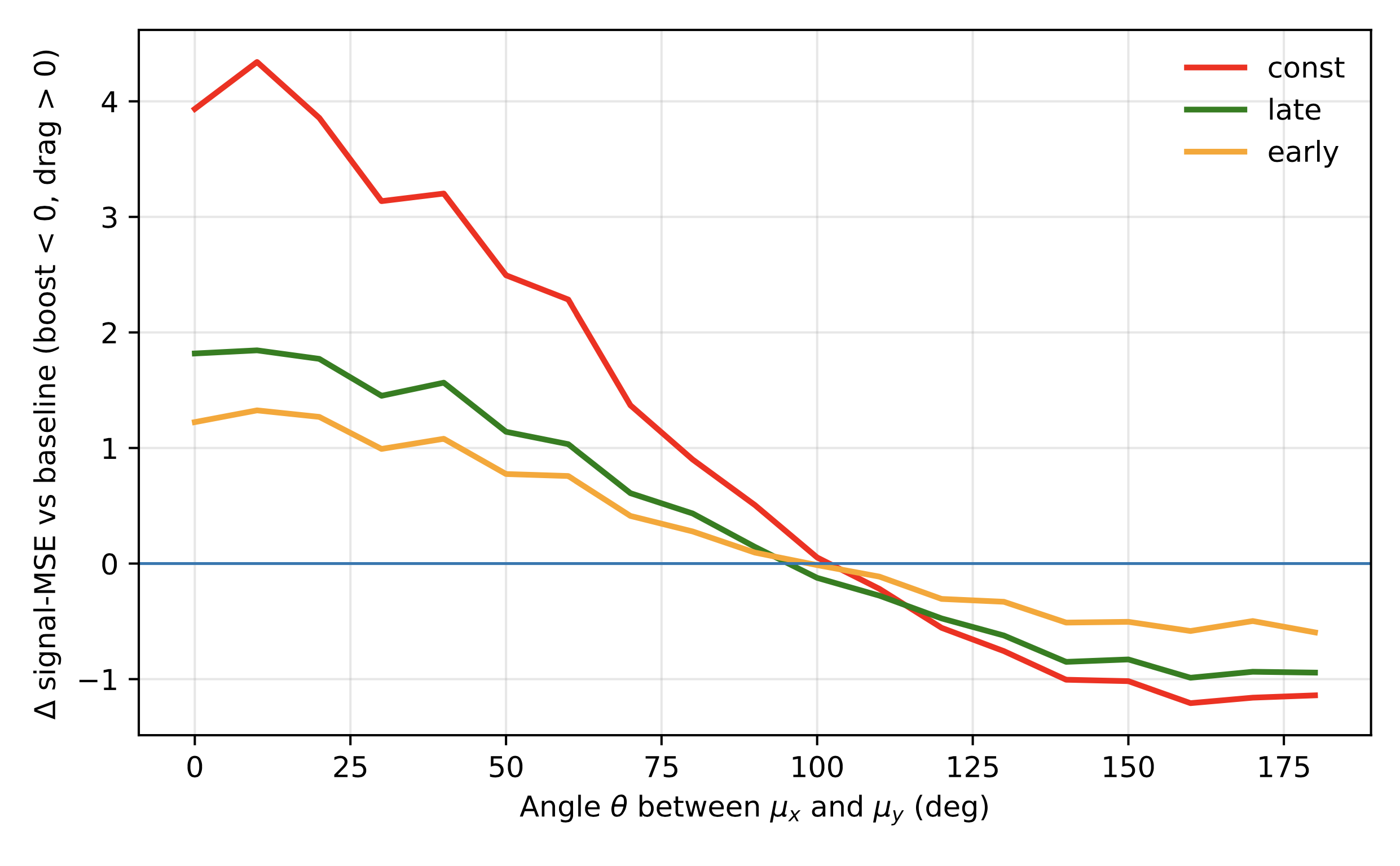}
    \caption{$\Delta$ MSE vs to baseline.}
    \end{subfigure}
    \centering
    \begin{subfigure}[b]{0.32\textwidth}
    \centering
    \includegraphics[width=\textwidth]{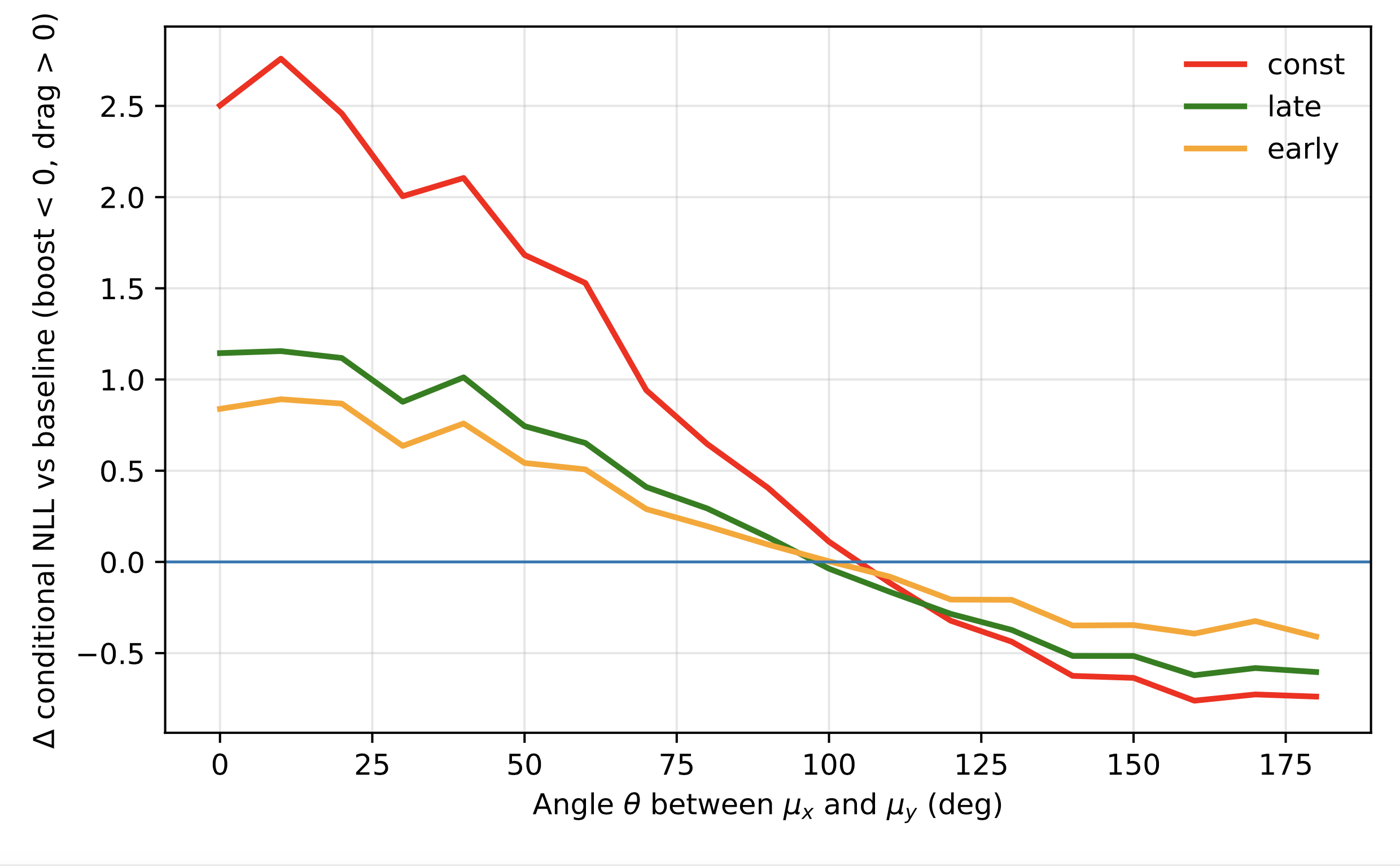}
    \caption{$\Delta$ NLL vs to baseline.}
    \end{subfigure}
    \caption{Results of the OU experiment: $\Delta$ of observable wrt. to baseline for a coupling strength $g_0=0.5$}
\end{figure}
\begin{figure}[ht]
    \centering
    \begin{subfigure}[b]{0.32\textwidth}
    \centering
    \includegraphics[width=\textwidth]{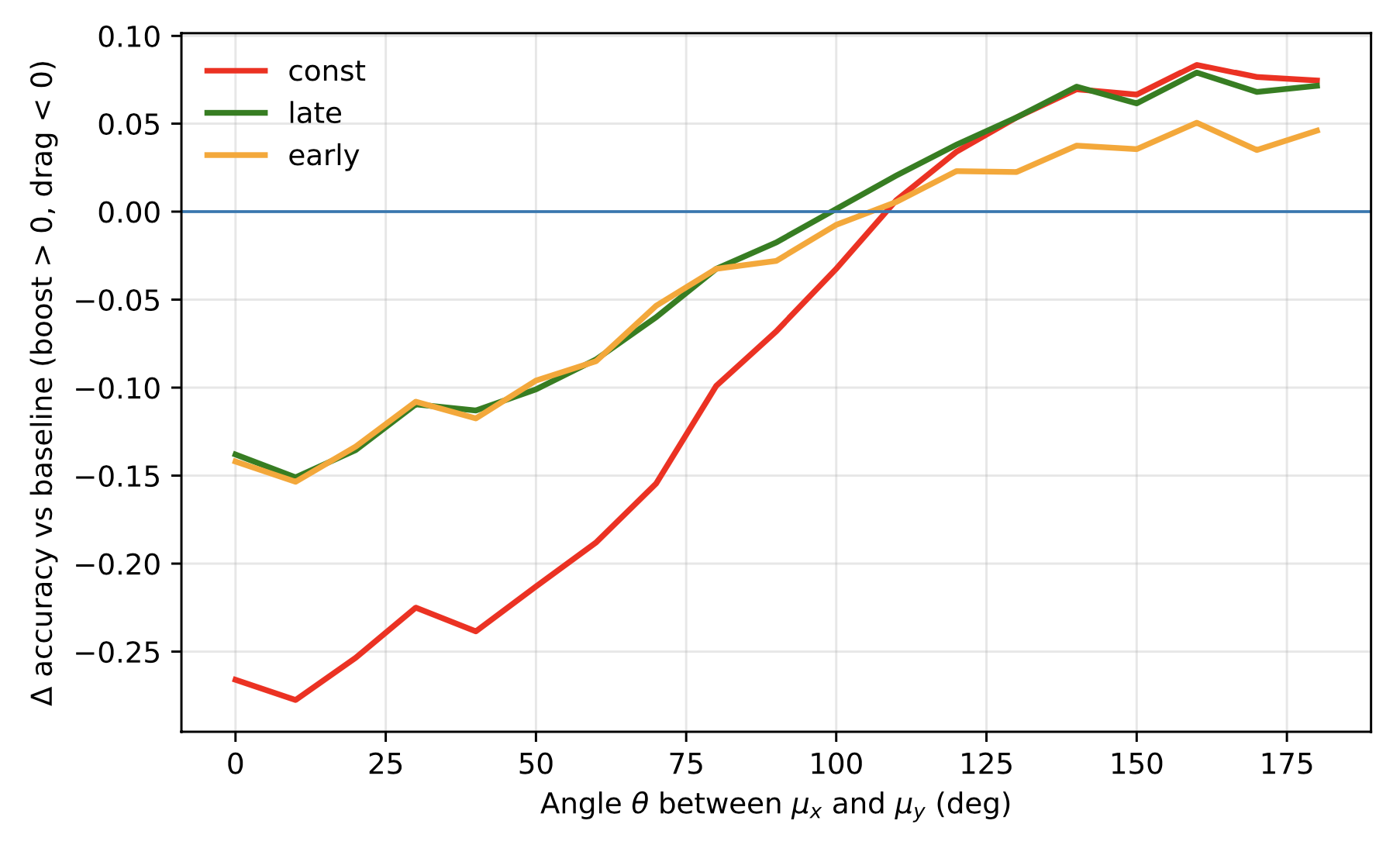}
    \caption{$\Delta$ accuracy vs to baseline.}
    \end{subfigure}
    \centering
    \begin{subfigure}[b]{0.32\textwidth}
    \centering
    \includegraphics[width=\textwidth]{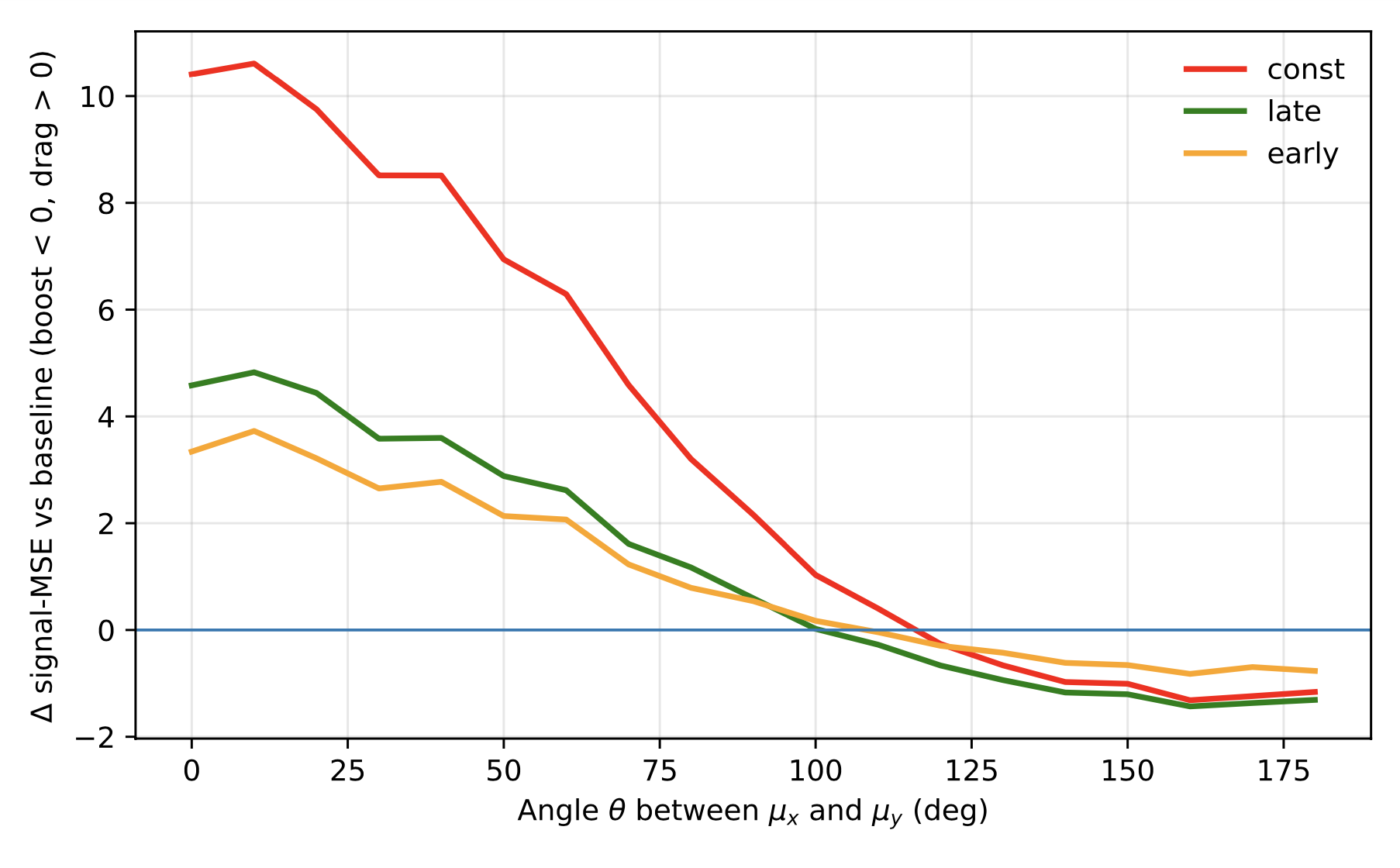}
    \caption{$\Delta$ MSE vs to baseline.}
    \end{subfigure}
    \centering
    \begin{subfigure}[b]{0.32\textwidth}
    \centering
    \includegraphics[width=\textwidth]{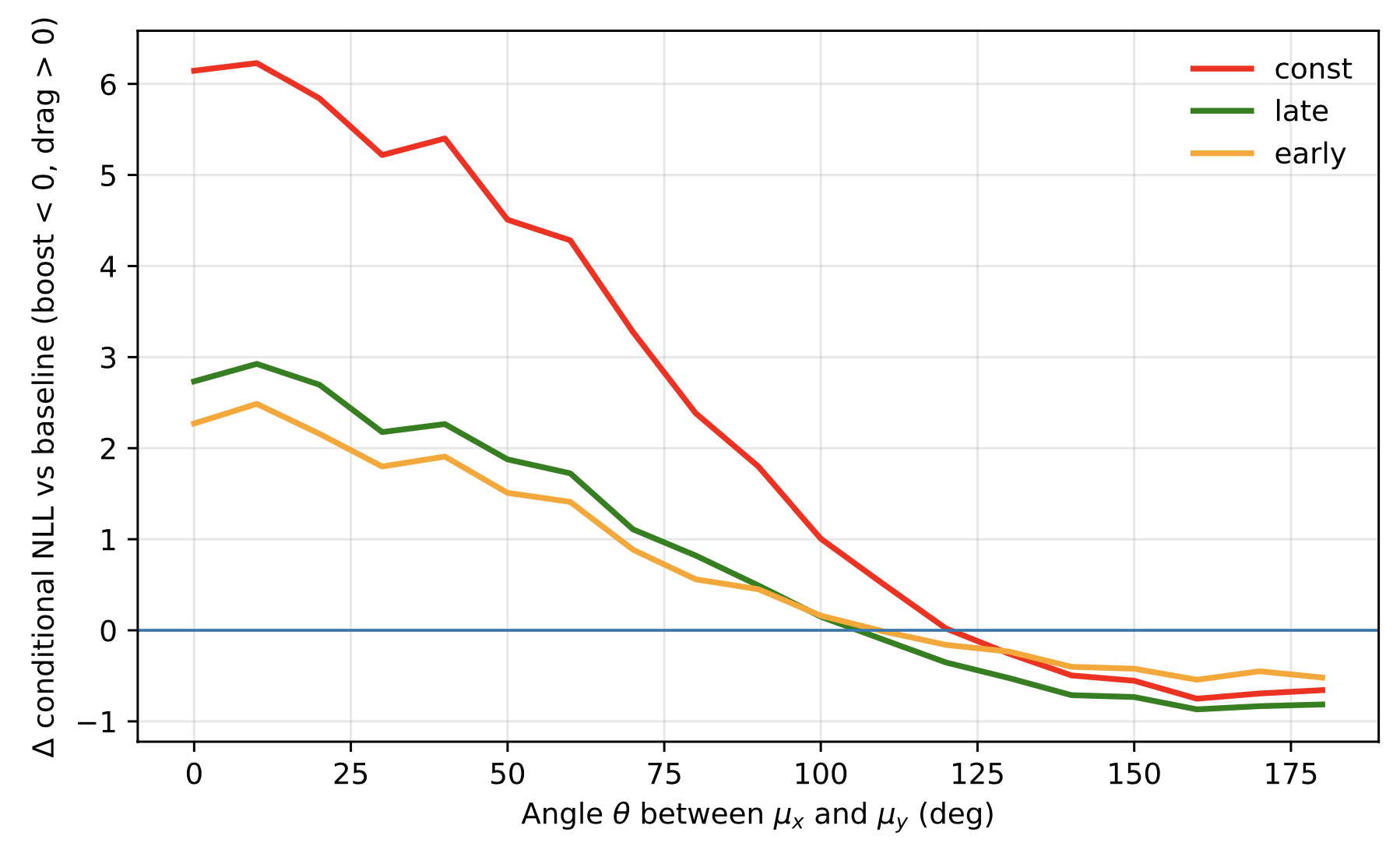}
    \caption{$\Delta$ NLL vs to baseline.}
    \end{subfigure}
    \caption{Results of the OU experiment: $\Delta$ of observable wrt. to baseline for a coupling strength $g_0=1.0$}
\end{figure}

\bibliographystyle{JHEP}
\bibliography{diff_ref.bib}

\end{document}